\documentclass[sigconf,authorversion, nonacm]{acmart}
\usepackage{subfigure}
\usepackage{multirow}
\usepackage[ruled,linesnumbered]{algorithm2e}

\DeclareMathOperator*{\argmax}{argmax}
\DeclareMathOperator*{\argmin}{argmin}

\AtBeginDocument{%
  \providecommand\BibTeX{{%
    \normalfont B\kern-0.5em{\scshape i\kern-0.25em b}\kern-0.8em\TeX}}}

\setcopyright{none}

\usepackage{ifthen}
\newboolean{final}
\setboolean{final}{false}

\def\filmroll{\textit{FacialFilmroll}\xspace}

\acmBooktitle{none}
\acmPrice{none}
\acmDOI{none}
\acmISBN{none}

\begin{document}

\title{\filmroll: High-resolution multi-shot video editing}


\author{Bharath Bhushan Damodaran}
\email{bharath.damodaran@interdigital.com}

\author{Emmanuel Jolly}
\affiliation{%
  \institution{InterDigital R\&I}
   \country{France}
}

\author{Gilles Puy}
\authornote{Contributed to this work while working at Interdigital R\&I, France, in 2019.}
\affiliation{%
  \institution{In his own name}
  \country{France}
}
\email{gilles.puy@gmail.com}

\author{Philippe Henri Gosselin }

\author{Cédric Thébault}

\author{Junghyun Ahn}
\affiliation{%
 \institution{InterDigital R\&I}
   \country{France}
  }
\email{{firstname.lastname}@interdigital.com}  

\author{Tim Christensen }

\author{Paul Ghezzo }
\affiliation{%
 \institution{In his own name}
   \country{USA}
  }

\author{Pierre Hellier}
\affiliation{%
 \institution{InterDigital R\&I}
   \country{France}
  }
\email{pierre.hellier@interdigital.com}

\renewcommand{\shortauthors}{Bharath Bhushan Damodaran, et al.}


\begin{abstract}
We present \filmroll, a solution for spatially and temporally consistent editing of faces in one or multiple shots. We build upon unwrap mosaic \cite{rav-acha2008unwrap} by specializing it to faces. We leverage recent techniques to fit a 3D face model on monocular videos to (i) improve the quality of the mosaic for edition and (ii) permit the automatic transfer of edits from one shot to other shots of the same actor. We explain how \filmroll is integrated in post-production facility. Finally, we present video editing results using \filmroll on high resolution videos.


\end{abstract}

%


\maketitle


%
\section{Introduction}
Video editing is one of the most demanding, challenging and labour-intensive tasks in the film post-production phase, where it takes several hours to edit a video of few seconds, even for senior artists. 
Almost all shots of an actor are retouched in some ways by an artists during this phase with, e.g., removal of eye-bags, rehab of jaw line, adding or removing scars or a moustache. Even if a makeup has been performed on-set, there is often a need for correction with digital makeup. 
One can even think of more advanced level of editing where the artists is asked to make the actor looks younger/older or add a smile.
Such edits need to be of extremely high quality as our brain is extremely skilled at detecting even slight modifications made on human faces, especially on high resolution videos.
A lot of time is thus spent by the artists to ensure that the edits are accurate and realistic.
In addition, digital makeup for a given actor is rarely done on a single shot. Indeed, the editing of a face should be made consistent across different shots, {making the task even more complex for an artist.}
To the best of our knowledge, there is currently no method that proposes {automatic} multi-shot editing{, which is the task that we address in this work.}

Given a video sequence, edits are conventionally performed on one frame and {propagated frame by frame to the rest of the shot.} 
The effectiveness of this approach depends on the quality of the spatio-temporal consistency of the edit propagation, which is directly dependent on the robustness of the forward and backward tracking methods \cite{yilmaz2006object,marvasti2021deep}. 
These {frame by frame} tracking methods often fails when the shots contain long range correspondences, object occlusions and disocclusions, extreme pose changes, or rapid motions, which are cases often observe in real-world shots. These cases require human corrections to ensure the spatio-temporal consistency of the edits in a shot. 
To tackle these limitations, \citet{rav-acha2008unwrap} introduced a method called \textit{unwrap mosaic}, where {an object in a} video is unfolded in a 2D texture image and the editing is performed on this texture map. The edited video is obtained by inverse projection of the texture map onto the image space using a bi-directional $2D$-$2D$ mapping and binary visibility masks. The bi-directional mapping allows us to transfer edits from the texture map to the image space and vice-versa. The binary visibility masks model the occlusions, indicating which parts of the texture map are visible in the image. 
{The advantage of this representation is twofold. First, it offers to the artists a greater flexibility than the conventional method as they can edit the entirety of the object on this mosaic; they are not limited to the partial views of the object in the original video. Second, long range correspondences can, theoretically, be easier to handle than in the conventional approach as edit propagation is done from a single representation of the whole object (the mosaic) to each frame via the $2D$-$2D$ mapping and not frame by frame where parts of the object are regularly (dis)occluded. Yet, in practice, the quality of this $2D$-$2D$ mapping is itself dependent of the quality of frame-by-frame tracking methods. Indeed, the first step to obtain this $2D$-$2D$ mapping is to track several points on the object across frames using, typically, the optical flow between frames.}

\begin{figure*}[t]
    \centering
    \includegraphics[scale=0.55]{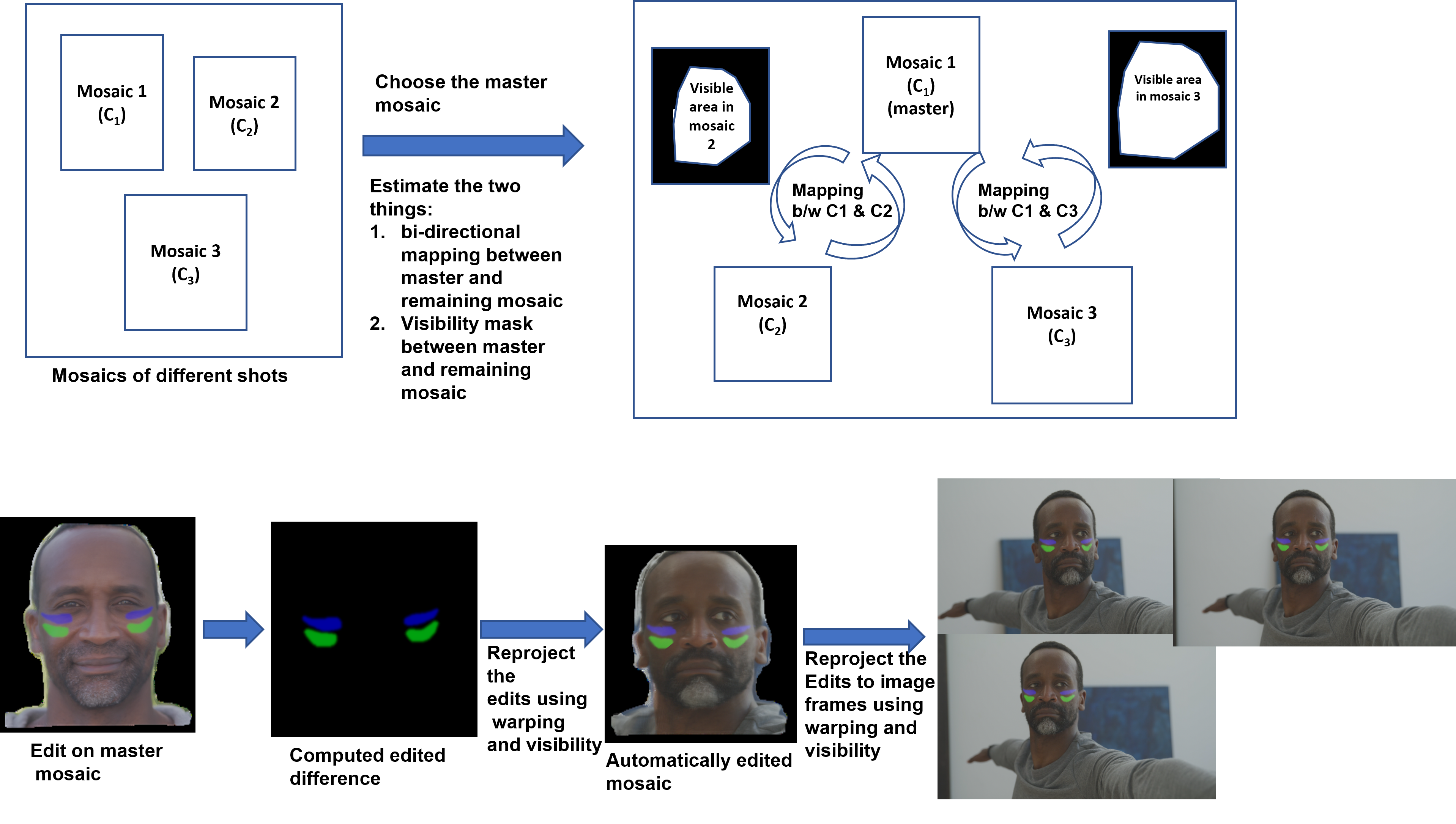}
    \caption{Illustration of multi-shot video editing. On top it illustrates the components in the multi-shot editing. On bottom it illustrates how the multi-shot editing is performed. The editing is performed on the master mosaic, and it is automatically transferred onto another mosaic using the warping and visibility between the mosaics. From the automatically edited mosaic, the edits are propagated to the frames using the warping and visibility estimated for the single-shot.}
    \label{fig:multishot-scheme}
\end{figure*}

In this work, we overcome this limitation of unwrap mosaic by leveraging the fact that we target edits on human faces. We call our proposed method \filmroll.
First, we propose to use face capture tracks which are extracted from 3D face extraction techniques \cite{andrus2020facelab}, allowing us to capture long-range correspondences by exploiting the known 3D geometry of human faces.
Second, we propose an hybrid embedding method that mixes face tracks and optical flow tracks in the estimation of the $2D$-$2D$ mapping between the mosaic and the video frames.
Third, we address the computational issues when using \filmroll on high-resolution videos by proposing a multi-resolution strategy where certain steps are only performed at the lowest resolutions.
Last, but not least, we propose a multi-shot version of \filmroll that allows an artist to perform an edit on one mosaic and propagate it consistently in multiple shots. The illustration of multi-shot approach is shown in figure~\ref{fig:multishot-scheme}. 
We demonstrate the potential of our proposed method in real-world conditions encountered in post-production environments.

The rest of the paper is organized as follows. Section \ref{sec:unwrap-model} introduces the \textit{unwarp mosaic} model. Section \ref{sec:filmroll} describes the \filmroll model and the details of its estimation. Section \ref{sec:multi-shot} describes our proposed multi-shot video editing. Section \ref{sec:workflow} describes how our method is used in the post-production settings. The experimental results are discussed in section \ref{sec:exp results} and we conclude in section \ref{sec:con}.

%
\section{Related work}

To the best of our knowledge, there has been no specific method to deal with the difficult problem of multishot editing. Current method relies on the spatio-temporal propagation of edits using optical flow or tracking method. We refer the reader to recent survey papers \cite{fortun2015optical,smeulders2013visual} on these methods. The consistency of editing on a single shot has been mainly addressed in the unwrap mosaic paper \cite{rav-acha2008unwrap} that we describe extensively in the next section.
The problem of multi-shot editing is more challenging since frames between two shots can differ in many ways with large displacements. To this aim, recent methods propose semantic alignment using deep features extraction and matching \cite{rocco2018end}. Unfortunately, none of these methods can achieve a robust spatial and temporal consistency across a set of frames taken from different shots. Further, capturing the global structure of the data in the lower dimension is crucial for our method. There are several embedding methods in the literature \cite{Maatentsne, Ngspectralemb, Roweis2323LLE, mcinnes2020umap, Adityanode2vec}, but they are limited in capturing global structure of the data \cite{mcinnes2020umap}. Thus our embedding method will be based on the multi-dimensional scaling which preserves the distance of the data in the embedded space.

%
\section{Unwrap Mosaic Model}\label{sec:unwrap-model}
In this section, we introduce the unwrap mosaic model \cite{rav-acha2008unwrap}, on which our \filmroll model is built upon. The unwrap mosaic model introduces a new image generation model from which sequences of images are formed. It assumes that the entire sequence of images of an object can be represented as the 2D texture map by parameterizing the object surface on the unit 2D square, and that the sequence can be reconstructed back from this 2D texture map.

\subsection{Continuous model}

Let $\{\mathbf{I(x}, t) \in \mathbb{R}^3 \}_{t=1}^{T}$ be a sequence of images of the surface of an object, where $\mathbf{x}=(x,y)$ are the coordinates in the image domain and $t$ is the temporal dimension. 
The unwrap mosaic model assumes that these images can be rendered from a unique texture map, called mosaic, represented as a function $\mathbf{C}: \mathbf{Q} \rightarrow \mathbb{R}^3$, where $\mathbf{Q}$ denotes the unit square on which the texture map is defined. 
One can think of this mosaic as a form \textit{uv} map representation of the 3D surface of the object but where the precise 3D geometry of this surface is nevertheless unknown.
An image $\mathbf{I}(\mathbf{x}, t)$ from this sequence can rendered from the mosaic $C$ thanks to a time-dependent $2D-2D$ mapping $\mathbf{w}: \mathbf{Q} \times \mathbb{R} \rightarrow \mathbb{R}^2$ from the unit square $\mathbf{Q}$ to the image coordinates via the formula:
\begin{equation}\label{eq2}
    \mathbf{\hat{I}(x}, t) = \int \rho(\mathbf{w(u},t) - \mathbf{x)} \, \mathbf{C(u)} \, J(\mathbf{u},t) \, d\mathbf{u},
\end{equation}
%
%
where $\rho$ denotes the point spread function (PSF) of the camera
and $J(\mathbf{u})$ is the determinant of the mapping Jacobian. Ideally, we should have $\mathbf{I(x}, t) = \mathbf{\hat{I}(x}, t)$. The texture map $\mathbf{C}$ is independent of $t$ as it is assumed that the object's color will remain the same for each frame of the video. Note that this model assumes, for the moment, that every point in $\mathbf{C}$ is visible in $\mathbf{\hat{I}(x)}$. 


In practice, parts of $\mathbf{C}$ may not be visible in every frames due to occlusions by other objects in the scene or even self-occlusion. The authors of the unwrap mosaic model handle this scenario by introducing a time-dependent binary visibility map $\mathbf{b}\mathbf{(u},t)$ (where $\mathbf{b}\mathbf{(u},t) = 1$ if the object surface at position $\mathbf{u}$ is visible in the $t^{\rm th}$ frame) in equation \eqref{eq2}, which becomes
\begin{equation}\label{eq:unwrapmodel}
    \mathbf{\hat{I}(x}, t) = \int \rho(\mathbf{w(u},t) - \mathbf{x)} \, \mathbf{C(u)} \, \mathbf{b}\mathbf{(u,}t) \, J(\mathbf{u},t) \, d\mathbf{u}.
\end{equation}
Equation \ref{eq:unwrapmodel} is called the \textit{unwrap mosaic} model. 

Finally, when multiple ($L$) objects appear in the video, the unwrap mosaic model becomes
\begin{equation}\label{eq:4}
    \mathbf{\hat{I}(x}, t) = \sum_{\ell=1}^L\int \rho(\mathbf{w}^\ell(\mathbf{u},t) - \mathbf{x)} \,\mathbf{C}^\ell\mathbf{(u)}\,\mathbf{b}^\ell\mathbf{(u,}t)\,J^\ell(\mathbf{u},t)\,d\mathbf{u},
\end{equation}
where each object is represented by a triplet $(\mathbf{C}^\ell,\mathbf{w}^\ell, \mathbf{b}^\ell)$ where $\mathbf{b}^\ell$ encodes inter-object occlusions and self-occlusions. Note that we are given only the image sequences $\mathbf{I(x,}t)$ as input and all the elements in this triplet are unknown.

\subsection{Discrete model}

In order to estimate the unknowns $\mathbf{C},\mathbf{w}, \mathbf{b}$ in \eqref{eq:unwrapmodel}, the \textit{unwrap mosaic} model is discretized, yielding a discrete ill-posed inverse problem solved by minimizing the sum of a data cost term and regularisation terms. 

Denoting, as in \cite{rav-acha2008unwrap}, the grid size of the parameter space $\mathbf{Q}$ as $w \times h$ and of the image space as $W \times H$, the number of variables to estimate is $w \times h \times 3$, for the texture map $\mathbf{C(u)}$, plus $w \times h \times T \times 2$, for the maps $\mathbf{w(u},t)$, plus $w \times h \times T$, for the visibility maps $\mathbf{b(u,}t)$.
The discretized version of \eqref{eq:unwrapmodel} is written as
\begin{equation}\label{eq:unwrapmodel-discrete}
    \mathbf{\hat{I}(x}, t) = \frac{\sum_u A(\mathbf{u, x},t) \mathbf{C(u)}b\mathbf{(u,t)}}
    {\sum_u A(\mathbf{u, x},t) b\mathbf{(u,t)}},
\end{equation}
where $A(\mathbf{u, x},t)$ is function of the mapping $\mathbf{w}$, its Jacobian, and the PSF $\rho$.

To avoid too much difference of resolution between the images  $\mathbf{\hat{I}(x}, t)$ and the mosaic $\mathbf{C(u)}$, one should choose a grid size $w \times h$ approximately equal to $W \times H$, indicating that the complexity of the model increases with the resolution of the image sequence.

\subsection{Estimation of parameters}
To estimate the parameters $\mathbf{C},\mathbf{w}$, and $\mathbf{b}$, \cite{rav-acha2008unwrap} minimises the energy 
\begin{equation}\label{eq:unwrap-enegery}
    E(\mathbf{C,w},b) =  E_{\rm data}(\mathbf{C,w},b) + \lambda_w E_w(\mathbf{w}) + \lambda_b E_b(b).
\end{equation}
where 
%
%
is the data fidelity term whose role is to ensure that $\mathbf{I(x},t) \approx \mathbf{\hat{I}(x},t)$, and  $E_w(\mathbf{w})$ and $E_b(\mathbf{b})$ are regularisation terms on the $2D$-$2D$ mapping and visibility maps. Since the resulting optimization problem is non-convex, \cite{rav-acha2008unwrap} use coordinate descent to minimize the total energy $E$, by alternating between estimation of $\mathbf{C},\mathbf{w}$, and $\mathbf{b}$. Please refer to \cite{rav-acha2008unwrap}, for the exact definition of $E_w(\mathbf{w})$ and $E_b(\mathbf{b})$ used in unwrap mosaic, as well as for the way the minimisation problem is solved. We will detail here the modification we brought to this cost function and our method to minimize it. 


\subsection{Limitations}
%
{The key to reach good editing quality with the \textit{unwrap mosaic} model to have a good mapping $\mathbf{w}$ between the mosaic space and the image space. This mapping is estimated by tracking points at the surface of the object of interest in the video sequence. One can use, e.g., the optical flows to track such points, which has the advantage of being agnostic to the type of surface. The drawback is that this approach usually fails to capture long term correspondences. Leveraging the fact that we are mostly interested in editing faces, we propose to complement optical flow tracks with tracks extracted using a 3D model of a face.}


%
\section{\filmroll}\label{sec:filmroll}
In this section, we describe \filmroll which consists of four main steps. First, the object of interest is segmented on each frame of the video sequence. Second, points at the surface of this object are tracked throughout the sequence. Third, these tracks are embedded in $\mathbf{Q}$, providing correspondences encoded by $\mathbf{w}$ between the mosaic space and the image space. Fourth, the mosaic $\mathbf{C}$ and visibility maps $\mathbf{b}$ are estimated. At the end of this process, any edits done by an artist on $\mathbf{C}$ can be propagated to the video frames thanks to the mapping $\mathbf{w}$.

The image formation model in \filmroll is identical to the unwrap mosaic one and we estimate its parameters by minimising a cost function which is similar to \eqref{eq:unwrap-enegery} and which takes the form
\begin{equation}\label{eq:filmroll-energy}
    E(\mathbf{C,w},b) =  E_{\rm data}(\mathbf{C,w},\mathbf{b}) + \lambda_b E_b(\mathbf{b}) + \lambda_c E_c(\mathbf{C}).
\end{equation}
We however made different choices of data fidelity and regularisation terms than in unwrap mosaic. We do not claim an advantage of these choices over the original costs beyond an easy minimisation of all of them by gradient descent. We choose
\begin{align}
    E_{\rm data}(\mathbf{C,w},b) &= \sum_t\sum_{\mathbf{x}} \Vert \mathbf{I(x},t) - \mathbf{\hat{I}(x},t) \Vert_1,
\end{align}
for robustness to pixel outliers, e.g., in presence of occlusions, as data fidelity term and
\begin{align}
    E_b(\mathbf{b}) = \sum_t\sum_\mathbf{x} \Vert \nabla_{\mathbf{x}} \mathbf{b}(\mathbf{x}, t) \Vert_2^2, 
    \text{ and }  
    E_c(\mathbf{C}) = \sum_\mathbf{x} \Vert \mathbf{C}(\mathbf{x}) \Vert_2^2,
\end{align}
the Tikhonov and $\ell_2$-norm regularizations, respectively. More details about the minimisation of this cost function $E$ are given in section \ref{sec:estoffilmroll}.

Unlike in unwrap mosaic, we did not include a cost $E_w(\cdot)$ in \eqref{eq:filmroll-energy} as we estimate the mapping $\mathbf{w}$ independently of $\mathbf{C}$ and $\mathbf{b}$ (see section \ref{sec:mapping}), or use off-the-shelf optical flow algorithms to refine this mapping. Nevertheless, the preliminary step to compute this mapping remains the tracking of points at the surface of the object of interest, as in \cite{rav-acha2008unwrap}, followed by an embedding of these tracks into $\mathbf{Q}$. In this work, we propose to leverage the fact that the objects of interest are faces and use two types of tracks: the first are obtained by using the optical flow between images, computed using \cite{Weinzaepfeldeepflow}, and permits us to track local deformations of faces, while the second are obtained by face capture and permits us to track the global geometry of the face. We describe in the next section how the face tracks are obtained. 
\subsection{Specialization to faces.}
To automatize the full process as most as possible, we use a method for multi-face tube extraction followed by an automatic segmentation technique designed for faces to isolate all the faces in a sequence.

\subsubsection{Multi face tube extraction}

In this step, we process a video sequence with possibly several faces appearing in it and output a set of face ``tubes'', where a tube is a contiguous sequence of 2D boxes containing the face of one person.

First, we detect all faces in each video frame independently using a neural network trained for multi-scale face detection \cite{zhang17iccv}. This network infers, for each frame and for each detected face, several candidate face boxes, from which we aim at building face tubes.
For the first frame, we select for each detected face, the face box with the best detection score. And for the other frames, we use the Hungarian algorithm to match the face box predicted by a Kalman filter with the candidate face boxes. Each matching box is added to the corresponding face tube. For each non-matched detected face, we create a new face tube. And we close non-matched face tubes.


\subsubsection{Automatic face delineation}

To restrict the computation of the \filmroll to faces only, an automatic semantic segmentation approach is used to segment the \textit{person} class from the background \cite{wu2019detectron2,kirillov2020pointrend}. This segmentation is more accurate than the bounding box face tube detection, since all parts of interest (including neck, hair, forehead) are included for potential editing. This allows to efficiently separate the person's silhouette and hair from the background.

\subsubsection{Face tracks}\label{subsec:tracks}

We use an inverse rendering method similar to \cite{garrido2016mono}~\cite{andrus2020facelab}~\cite{dib2021drt} to fit a 3D mesh on each face tube.
This model consists in parameters describing the facial shape and albedo using a 3D morphable model \cite{blanz1999mm} \cite{egger2020mmfuture}, the facial expression (blendshapes), the head pose (rotation and translation) and the illumination (spherical harmonics). The parameters of this model are estimated as follows.
\paragraph{Coarse modelization:}
The pose and the expression parameters are initialized using facial landmarks, inferred by a neural network \cite{bulat17far}. This initialization is achieved by minimizing a loss that computes the distance between the facial landmarks predicted by the neural network and the projection of their assigned location on the 3D mesh.
\paragraph{Shape, color, and light learning:} We extract up to 60 frames from the primary face tube and compute a loss that compares the projection of each vertex and the corresponding pixel in the video frame. We use this loss to learn the shape, color, and light parameters and refine the pose and expression parameters. 
\paragraph{Fine modeling:} We process each frame of the primary face tube, from the first to the last, to learn per-frame parameters (blendshapes coefficients and head pose), using a similar approach as in the previous step. To ensure a temporal consistency, we penalize differences between the parameters of adjacent frames.
\paragraph{Final face tracks:} From this fitted 3D mesh, we extract the trajectories (the 2D projection) of the vertices along the sequence. The vertices which are not visible in a given frame are flagged as such.


\subsection{Hybrid Embedding}\label{subsec:hybrid-emb}
In this subsection, we describe our construction of the hybrid embedding function $\phi(\cdot)$ that maps the optical flow and face capture tracks to $\mathbf{Q}$. We recall that the face capture tracks permit us to preserve the topological/global structure of the face in the mosaic, while the optical flow tracks permit to refine local geometric structures in the face.


Let $T$ be number of the frames in the video. We start by tracking points on the surface of the object of interest with optical flow, using the method of \cite{Weinzaepfeldeepflow}. This yields tracks $\{\mathbf{x}_i(t) \in \mathbb{R}^2\}_{t=1}^T$ where $i$ index the tracked point in the optical flow method. We also track points using face capture \cite{andrus2020facelab}, which yields tracks $\{\mathbf{z}_i(t) \in \mathbb{R}^2\}_{t=1}^T$. Note that $\mathbf{x}_i(t)$, and $\mathbf{z}_i(t)$ do not correspond to the same physical point on the object. Following the embedding principle in \cite{rav-acha2008unwrap}, $\phi$ is constructed to preserve the maximum distance between each pair of points observed along the tracks. Hence, we define
\begin{align}
    d_{ij}^{(1)} &= \mathbf{x}_i(t) - \mathbf{x}_j(t), \text{ where} \ 
    t = \argmax_{\ell \in T_i^x \cap T_j^x}  \Vert \mathbf{x}_i(\ell) - \mathbf{x}_j(\ell) \Vert_2, \label{eq:maxdist_opt}
    \\
    d_{ij}^{(2)} &= \mathbf{z}_i(t) - \mathbf{z}_j(t), \text{ where} \
    t = \argmax_{\ell \in T_i^z \cap T_j^z}  \Vert \mathbf{z}_i(\ell) - \mathbf{z}_j(\ell) \Vert_2,
    \label{eq:maxdist_fc}
\end{align}
where ${T_i^x}$ and ${T_j^x}$ are the set of time indices where point $i$ and $j$ tracked by optimal flow are visible, and where ${T_i^z}$ and ${T_j^z}$ are the set of time indices where point $i$ and $j$ tracked by face capture are visible. We also define 
\begin{equation}
    d_{ij}^{(12)} = \mathbf{x}_i(t) - \mathbf{z}_j(t), \text{ where} \
    t = \argmax_{\ell \in T_i^x \cap T_j^z}  \Vert \mathbf{x}_i(\ell) - \mathbf{z}_j(\ell) \Vert,
    \label{eq:maxdist_btw}
\end{equation}
which measure the distance between two points, each tracked with a different method.

The embedding function $\phi(\cdot)$ takes as input a track $\{\mathbf{x}_i(t)\}_{t=1}^T$ or a face capture track $\{\mathbf{z}_i(t)\}_{t=1}^T$ and yields a corresponding point in $\mathbf{Q}$ denoted by
\begin{align}
\phi(\{\mathbf{x}_i(t)\}_{t=1}^T) = \mathbf{u}_i^{(1)}
\text{ and }
\phi(\{\mathbf{z}_i(t)\}_{t=1}^T) = \mathbf{u}_i^{(2)},
\end{align}
respectively. This function is constructed to preserve, as well as possible, the relative positive between tracked points when these ones are maximally separated in the image sequence. Hence, the goal is to have 
\begin{align*}
d_{ij}^{(1)} \approx \mathbf{u}_i^{(1)} - \mathbf{u}_j^{(1)}, \    
d_{ij}^{(2)} \approx \mathbf{u}_i^{(2)} - \mathbf{u}_j^{(2)}, \text{ and }    
d_{ij}^{(12)} \approx \mathbf{u}_i^{(1)} - \mathbf{u}_j^{(2)}.
\end{align*}
The embedding function $\phi(\cdot)$ is obtained by solving an optimisation problem built to satisfy these constraints as well as possible. We solve the same optimisation problem as in \cite{rav-acha2008unwrap} whose objective function becomes
\begin{align}\label{eq:hybrid-obj}
L(\mathbf{u}^{(1)}, \mathbf{u}^{(2)}) 
= 
& \sum_{k,i,j}
    \exp( -{ \Vert d_{ij}^{(k)}\Vert_2^2}/{\tau} ) \;
    \Vert d_{ij}^{(k)} - (\mathbf{u}_i^{(k)} - \mathbf{u}_j^{(k)}) \Vert_2^2  
\nonumber\\ 
& + 2 \, \sum_{i,j} \exp( -{\Vert d_{ij}^{(12)}\Vert_2^2}/{\tau} ) \; \Vert d_{ij}^{(12)} - (\mathbf{u}_i^{(1)} - \mathbf{u}_j^{(2)}) \Vert_2^2, 
\end{align}
when considering both type of tracks. We have
\begin{align}\label{eq:hybrid-prob}
\mathbf{u}_1^{(1)}, \ldots, \mathbf{u}_N^{(1)}, \mathbf{u}_1^{(2)}, \ldots, \mathbf{u}_M^{(2)} = \argmin_{(\mathbf{v}^{(1)}, \mathbf{v}^{(2)})}  L(\mathbf{v}^{(1)}, \mathbf{v}^{(2)}),
\end{align}
where $N, M$ are the number of tracked points by optical flow and face capture, respectively. One can derive a closed form solution to Problem \eqref{eq:hybrid-prob} but we observed that this solution is not satisfactory as it is, in general, biased towards the preservation of the geometry captured by the optical flow tracks, making our effort to get face capture tracks vain. To avoid this issue, we propose the \emph{ad-hoc} but effective Algorithm \ref{alg:embedding}.
\begin{algorithm}[t]
\caption{Computation of the embedding $\phi$}
\small
$\{\mathbf{u}_i^{(2)}\}_{i=1}^M 
\gets 
\argmin_{\mathbf{v}^{(2)}}  
\sum_{i,j} \exp( -{d_{ij}^{(2)}}/{\tau} ) \; \Vert d_{ij}^{(2)} - (\mathbf{v}_i^{(2)} - \mathbf{v}_j^{(2)}) \Vert^2$\\
{$\mathbf{u}_i^{(1)} \gets \mathbf{0}$ for $i=1, \ldots, N$.}\\
\For{$\ell = 1, \ldots, \ell_{\rm max}$}{
    $\mathbf{u}_i^{(1)} \gets \mathbf{u}_i^{(1)} - \mu_1 \; \partial_{\mathbf{u}_i^{(1)}} L(\mathbf{u}^{(1)}, \mathbf{u}^{(2)})$\\
    \If{$\ell > \ell_{\rm min}$}{
    $\mathbf{u}_i^{(2)} \gets \mathbf{u}_i^{(2)} - \mu_2 \; \partial_{\mathbf{u}_i^{(2)}} L(\mathbf{u}^{(1)}, \mathbf{u}^{(2)})$
    }
}
\label{alg:embedding}
\end{algorithm}
Note that we start by initialising $\mathbf{u}_i^{(2)}$ using only the face capture tracks to make sure to preserve the geometry of the face in the mosaic. Then, we update the embedding of the optical flow tracks by gradient descent with a stepsize $\mu_1$ while keeping the face capture embedding $\mathbf{u}_i^{(2)}$ fixed during the first $\ell_{\rm min}$ iterations. During the last iterations, we also update the face capture embedding but with a stepsize close to zero. These last iterations permit to correct for the small tracking errors of the face capture method.

\subsection{Estimation of the \filmroll parameters} \label{sec:estoffilmroll}
In this subsection, we describe how we estimate the unknowns $\mathbf{w}, \mathbf{b}, \text{and} \mathbf{C}$ of the \filmroll model.

\subsubsection{Estimation of 2D-2D mapping $\mathbf{w}$}
\label{sec:mapping}

 The mapping $\mathbf{w}$ indicates how every pixel's $\mathbf{u}$ on the mosaic space is mapped in the image space. Thanks to the embedding $\phi$ of the tracks, we already know that we should have $\mathbf{w}(\mathbf{u}^{(1)}_i, t) = \mathbf{x}_i(t)$. To be able to compute $\mathbf{\hat{I}(x}, t)$ defined in \eqref{eq:unwrapmodel-discrete}, we however need the knowledge of $\mathbf{w}(\mathbf{u}_k, t)$ for every position $\mathbf{u}_k \in \mathbf{Q}$ on the grid of size $w \times h$. We obtain this information by interpolation between the known correspondences $\mathbf{w}(\mathbf{u}^{(1)}_i, t) = \mathbf{x}_i(t)$. Note that we only consider the embedding of the optical flow tracks in this step. The interpolation is computed using the following principle: if two points $\mathbf{u}$ and $\mathbf{v}$ are close in the mosaic space then their mapping $\mathbf{w}(\mathbf{u})$ and $\mathbf{w}(\mathbf{v})$ should also be close in the image space. 
Let $\mathbf{v}_k = \mathbf{u}_k$ for $k = 1, \ldots, wh$ and $\mathbf{v}_{wh+i} = \mathbf{u}^{(1)}_i$ for $i = 1, \ldots, N$. The mapped points satisfy $\mathbf{w}(\mathbf{u}_k, t) = \mathbf{x}^t_{k}$, $k = 1, \ldots, wh$, where $\mathbf{x}^t_{k}$ is the solution of
\begin{align}
\label{eq:subw-energy}
    & \min_{\mathbf{y}^t_{1}, \ldots, \mathbf{y}^t_{wh+N}} \; \sum_{k}\sum_{k' \in \mathcal{N}(k)} \exp(-\Vert \mathbf{v}_k - \mathbf{v}_{k'}  \Vert_2^2 / \tau) \; \Vert \mathbf{y}_k^t - \mathbf{y}_{k'}^t \Vert_2^2 \nonumber\\
    & \text{subject to } \quad \mathbf{y}^t_{wh+i} = \mathbf{w}(\mathbf{u}^{(1)}_i, t), \ i = 1,2, \dots, N,
\end{align}
where $t>0$, and $\mathcal{N}(i)$ is the set of $K$ nearest neighbors to $\mathbf{v}_k$ in $\{\mathbf{v}_i\}_{i=1}^{wh+N}$. Problem \eqref{eq:subw-energy} is solved by projected gradient descent using, e.g., an algorithm such as FISTA \cite{fista}. The resulting estimated mapping with this technique is denoted by $\mathbf{\hat{w}}$.

Let us mention that, in practice, we also estimate the inverse mapping $\mathbf{\hat{w}}^{-1}$ for a more efficient implementation of the algorithms. Once these mappings are available, the mosaic is initialised as $\mathbf{\hat{C}}(\mathbf{u}_k) = T^{-1} \sum_{t=1}^T \mathbf{I(\mathbf{w}(u_k)}, t)$.

\subsubsection{Estimation of visibility masks} 

Given the current estimation $\mathbf{\hat{w}}$ and $\mathbf{\hat{C}}$ of $\mathbf{w}$ and $\mathbf{C}$, respectively, we obtain a new estimate $\mathbf{\hat{b}}$ of the visibility masks by minimizing 
\begin{align}\label{eq:vis-energy}
    E_{\rm data}(\mathbf{\hat{C}, \hat{w}}, \mathbf{b}) + \lambda_b E_{b}(\mathbf{b})
\end{align}
over $\mathbf{b}$, while maintaining $\mathbf{\hat{w}}$ and $\mathbf{\hat{C}}$ fixed. The regularization term $E_{b}(\mathbf{b})$ is used to avoid sharp discontinuities in the visibility mask. Here we use the Tikhonov regularisation. This optimisation problem is separable in $t$ and each $\mathbf{b}(\cdot, t)$ can be computed in parallel. We estimate these visibility masks by gradient descent method using the Adam optimizer \cite{kingma2017adam}.

\subsubsection{Estimation of the texture map $\mathbf{C}$}
Given the current estimation $\mathbf{\hat{w}}$ and $\mathbf{\hat{b}}$ of $\mathbf{w}$ and $\mathbf{b}$, respectively, we obtain a new estimate $\mathbf{\hat{C}}$ of the mosaic by minimizing
\begin{align}\label{eq:mos-energy}
E_{\rm data}(\mathbf{{C}, \hat{w}}, \hat{b}) + \lambda_c E_{c}(\mathbf{C})
\end{align}
over $\mathbf{C}$, while maintaining $\mathbf{\hat{w}}$ and $\mathbf{\hat{b}}$ fixed. This optimisation problem is solved by stochastic gradient descent using the Adam optimizer \cite{kingma2017adam}, with batches selected over the frames.


\subsubsection{Mapping refinement and Lighting}
The 2D-2D mapping function $\mathbf{w}$ is the most important parameter which determines the accuracy of the visibility masks $b$, the mosaic $\mathbf{C}$, the reconstructed images $\mathbf{\hat{I}}(\mathbf{x}, t)$, and also the spatial and temporal consistency of the re-projected edits. To improve the overall accuracy, the mapping function $\mathbf{w}$ is refined after the first estimation of $\mathbf{C}$ and $\mathbf{b}$ by computing the optical flow between the reconstructed images $\mathbf{\hat{I}}(\mathbf{x}, t)$ and original image $\mathbf{{I}}(\mathbf{x}, t)$. Similarly, the lighting conditions also play a significant role in the quality of model. Therefore, the reconstructed image $\mathbf{\hat{I}}(\mathbf{x}, t)$ is corrected with a per-pixel affine correction with a strong smoothness regularisation on these parameters, just as done in section 3.5 of \cite{rav-acha2008unwrap}.

\subsection{Multi-resolution strategy}
The number of parameters to be estimated for the \filmroll increases dramatically with the content resolution. In particular, this can become an issue when dealing with $2K$ or $4K$ content, which is becoming common practice in a professional postproduction context.
To deal with this issue, we have set up a classical multiresolution strategy, composed of the following steps:
\begin{itemize}
    \item The input content is spatially downsampled by a factor $2^r$, where $r$ is chosen so that the number of pixels does not exceed a threshold value. Typically for $4K$ content, $r=3$ was used.
    \item At the coarsest resolution level, the tracks as well as the embedding is computed as described previously. Since the content resolution is reduced, that makes the computation of the embedding tractable in a reasonable time. Then, all variables  $\mathbf{C,w},\mathbf{b}$ are computed as described previously.
    \item We then iterate over the resolution levels until the finest resolution corresponding to the initial resolution. To do so, the previously computed variables $\mathbf{C,w},\mathbf{b}$ are upsampled from the previous resolution and the magnitude of the warps $\mathbf{w}$ is also multiplied by a factor two. Then, the variables are refined so as to minimize equation \eqref{eq:filmroll-energy} as described previously in section \ref{sec:estoffilmroll}.
\end{itemize}
In practice, this simple multiresolution strategy enables to handle up to $8K$ content using standard computation facilities, while limiting the amount of memory and cpu usage. It is noted multiresolution stratgey does not affect the visual quality or fidelity of the edits because the coarsest resolution level is only used as the initalization for the finest resolution variables.

\subsection{Multi-Shot mosaic construction}\label{sec:multi-shot}
In this section, we describe the adaptation of \filmroll for multi-shot video editing, where the objective is to perform editing jointly in different video shots of same actor. 

Let $\mathbf{C}_1, \mathbf{C}_2, \dots, \mathbf{C}_K$ be the mosaics of the $K$ shots which needs to be edited jointly. 
 Our proposal is to compute a universal mosaic $\mathbf{U}$ so that the correspondence from one mosaic $\mathbf{C}_i$ to all other one $\mathbf{C}_j$ can be found through the universal mosaic $\mathbf{U}$. The universal mosaic is computed thanks to a spatial embedding that preserve the geometric structure within each individual mosaic but also by ensuring that corresponding parts of the object between two mosaics have the same embedding. We denote $\mathbf{y}_i^{(k)}$ the $i^{\rm th}$ point sampled on the $k^{\rm th}$ mosaic and define
\begin{align}
    e_{ij}^{(k)} &= \mathbf{y}_i^{(k)} - \mathbf{y}_j^{(k)}.
\end{align}
We also compute correspondences between mosaic and define 
\begin{align}
    p_{ij}^{k\ell} &= \left\{ 
    \begin{array}{ll}
         1 & \text{if } \mathbf{y}_i^{(k)} \text{ and } \mathbf{y}_i^{(\ell)} \text{ are in correspondences,}\\
         0 & \text{oth.}
    \end{array}
    \right.
\end{align}
%
The correspondences between mosaics could be found using optical flow if the mosaics are not very different, or by matching keypoints. In our case, we leverage the available face capture tracks: each of the tracks corresponds to a unique vertex on the 3D face mesh, therefore points on the same vertex in two different mosaics are in correspondence. The embedding for the universal mosaic is the solution of
%
%
\begin{align}\label{eq:multi-shot}
\argmin_{\mathbf{u}_1^{(1)}, \ldots, \mathbf{u}_N^{(K)}}
& \sum_{k=1}^K \sum_{i,j}
    \exp( -{ \Vert e_{ij}^{(k)}\Vert_2^2}/{\tau} ) \;
    \Vert e_{ij}^{(k)} - (\mathbf{u}_i^{(k)} - \mathbf{u}_j^{(k)}) \Vert_2^2  
\nonumber\\ 
& +  \sum_{k,\ell=1}^K \sum_{i,j} p_{ij}^{k\ell} \; \Vert \mathbf{u}_i^{(k)} - \mathbf{u}_j^{(\ell)}) \Vert_2^2, 
\end{align}
The first term preserves the structure within each mosaic while the second ensure that two corresponding points in different mosaics have the same embedding. Problem \eqref{eq:multi-shot} is minimized by stochastic gradient descent. The universal mosaic is then computed using the same algorithm as for the individual mosaic.

%
\section{Workflow integration} \label{sec:workflow}
In this section, we discuss how \filmroll is integrated and used in a Post-Production facility. 
In order to control the computational burden and maximize the artist empowerment, \filmroll is used in two phases:
\begin{itemize}
    \item A batch processing is sent to a render farm for offline computation of the mosaic, warps and visibility masks, as described previously.
    \item Once this offline computation is finished, the artists can load the results in a dedicated Nuke plugin, named \textit{filmunroll}, to propagate edits done on the mosaic to the input video sequence.
\end{itemize}
We here describe how the two stages of the process have been developed.

\subsection{Batch processing}
The batch processing step uses a render farm at Post-Production facility where computers are isolated from the internet on a closed network. In order for \filmroll to be fully integrated in the post-production pipeline, we use containers\footnote{https://en.wikipedia.org/wiki/OS-level\_virtualization}. They allow us to easily define a custom execution environment (Operating System, Python version, \emph{etc}.), totally isolated from the host machine to not interfere with it, and very easy to replicate in different production sites.

The batch processing software comes with an API compatible with the production software Shotgun,\footnote{https://www.shotgunsoftware.com/}, and has been designed to ingest $4K$ exr\footnote{https://en.wikipedia.org/wiki/OpenEXR} content  on machines with $128GB$ of memory. The execution in \textbf{containers} allows us to control the memory and CPU usage. One hour of computation is currently needed for a standard footage ($2K$ content with $50$ frames) to extract the mosaic.

\subsection{Online editing in Nuke}
\begin{figure}[t]
    \centering
    \includegraphics[scale=0.6]{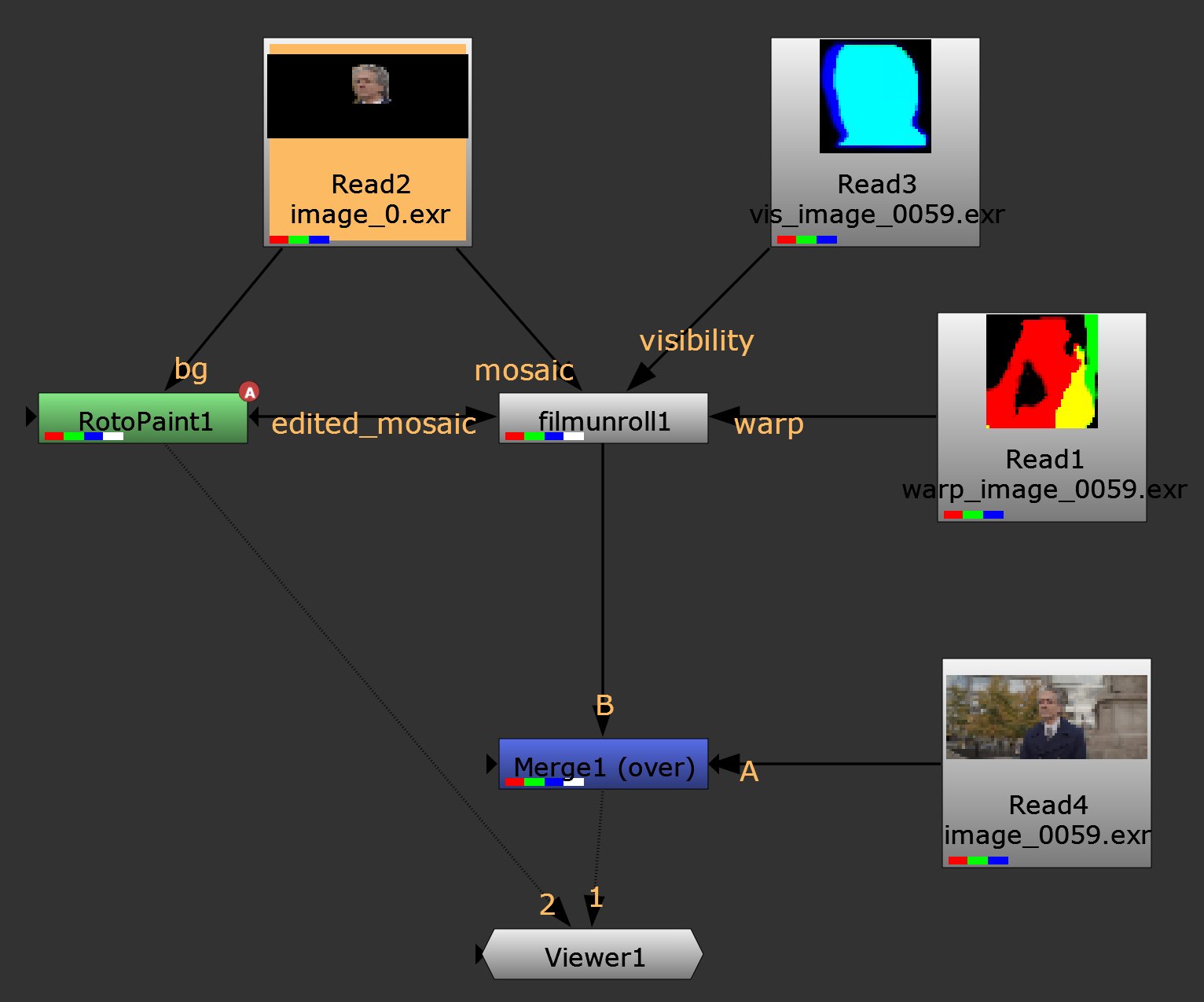}
    \caption{Illustration of the Nuke graph allowing the artists to use the pre-computed mosaic and project edits done on the original mosaic to each frame of the sequence.}
    \label{fig:nuke-reproj}
\end{figure}

Once the mosaic is computed, the artists can edit the mosaics in real time using Nuke with the \textit{filmunroll} plugin. \textit{Filmunroll} is an OpenFX\footnote{http://openfx.sourceforge.net/} plugin, which does not need to be recompiled for every new Nuke version (unlike Nuke native plugins).  
The \textit{filmunroll} plugin takes as inputs: the generated mosaic and its associated metadata, and the edited mosaic created by the artists. Instead of directly reconstructing the footage via modified mosaic, we actually compute the difference between the original and modified mosaics and project only this difference on each frame of the sequence.
 A Nuke graph allows the artist to superimpose this projected difference on the original shot using the Nuke \textit{Merge}\footnote{https://learn.foundry.com/nuke/13.0/content/reference\_guide/merge\_nodes/merge.html} node, as illustrated in figure \ref{fig:nuke-reproj}.

%
\section{Experimental Results} \label{sec:exp results}
We present here qualitative results obtained with \filmroll on high resolution videos via a visual inspection of the quality of the mosaic, the reconstructed images, and the re-projected edits. We also report computational time to compute the mosaic. 
We consider videos from the filmpac\footnote{https://filmpac.com/} website, where 
videos are available in both $HD$ and $4K$ resolution. For our experiments, we selected a set of $4K$ video sequences for both single-shot and multi-shot video editing. 

\subsection{Effect of our hybrid embedding}
\begin{figure}[t]
    \includegraphics[width=1.0\columnwidth]{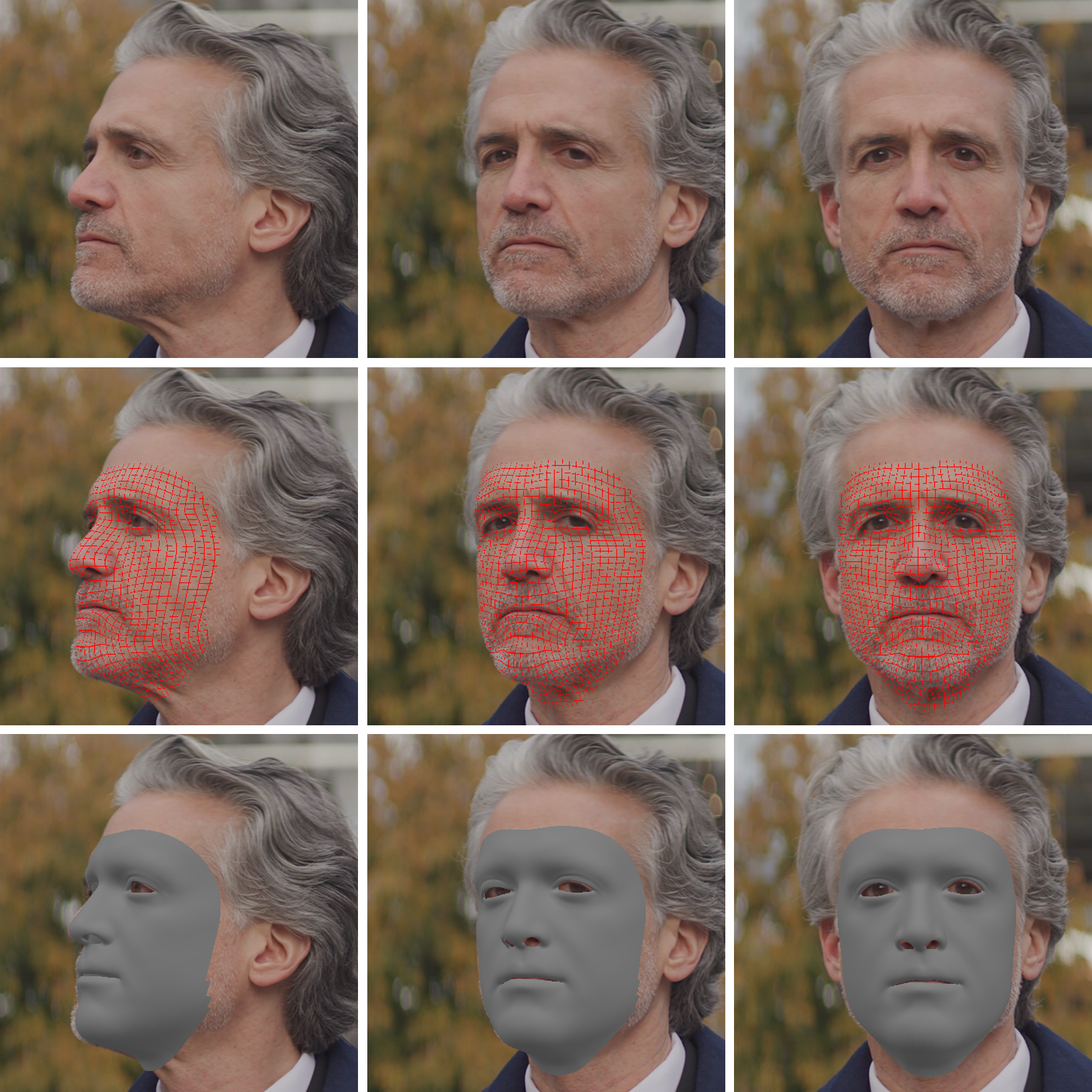}
    \caption{Few frames of the shot \textit{FP00673MD02} from the filmpac website. From left to right: frame 0, frame 50, and frame 100. Top: Original frames. Middle: 2D grid on our reconstructed 3D facial mesh. Bottom row: our reconstructed 3D facial mesh (see subsection \ref{subsec:tracks})}
    \label{fig:org_seq}
\end{figure}
%
{We assess here the impact of our proposed hybrid embedding on the quality of the reconstructed mosaic. We consider the shot \textit{FP00673MD02} in the filmpac website, which consists of $100$ frames, and where the actor turns his head from his right to his left while closing and opening his eyes in the middle of the sequence. The right part of his face is occluded in the first frames. Few frames of this sequence and the 3D reconstructed mesh are presented in Figure~\ref{fig:org_seq}.}
{The hybrid embedding is computed using $1000$ face tracks as well as the most reliable\footnote{Tracks which are consistent using both forward and backward tracking on the entire sequence.} optical flow tracks. We use $\ell_{\rm min} = 5$ and $\ell_{\rm max} = 10$ in our algorithm. We remarked that our hybrid embedding method converges quickly. }

\begin{figure}[ht] 
  \subfigure{%
    \includegraphics[scale=0.4 ]{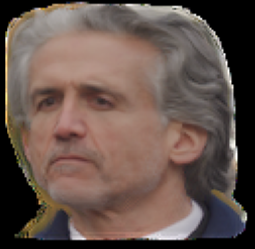}  
  } 
  \subfigure{%
    \includegraphics[scale=0.4 ]{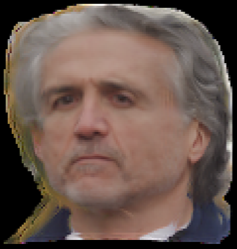}   
  } 
  \subfigure{%
    \includegraphics[scale=0.4 ]{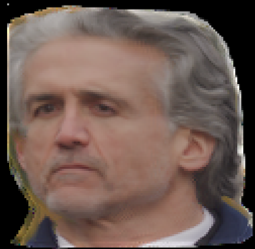}   
  } 
  \caption{Mosaics computed using only optical flow tracks (left), only face capture tracks (middle), and our hybrid embedding of optical flow and face capture tracks (right). The close inspection of the mosaic reveals the right side of the face and ear are better unfolded in the hybrid embedding.} 
  \label{fig:texturemap-emebdding}
\end{figure}

{We present in figure \ref{fig:texturemap-emebdding} the mosaics computed using the three different type of tracks: optical flow tracks alone, face capture tracks alone, our hybrid embedding with both optical and face capture tracks. First, we notice that all mosaics are of good quality. Second, we remark that the mosaic obtained with the face capture tracks has the most fronto-parallel pose with a better unfolding of the actor's right cheek than the mosaic obtained from the optical tracks alone. On the contrary, the mosaic obtained with the optical tracks presents a better unfolding of the actor's left ear. Our hybrid embedding achieves the best compromise with a good unfolding of both the left and right side of the actor's face, making it easier to edit both side of his face.}


\subsection{Multi-resolution framework} The advantage of the multi-resolution strategy is demonstrated in terms of computational time. For a sequence of size $2048 \times 1152$ with $50$ frames, \filmroll computed with our multi-resolution strategy is 3$\times$ faster ($3015$ seconds) than without mutli-resolution strategy ($9865$ seconds). 
The computational time is measured using a system with $64$ GB of RAM and a $12$ cores processor (Intel® Xeon® Gold 6128 CPU @ 3.40GHz).


\subsection{Single shot video editing}
\begin{figure*}[ht] 
  \subfigure{%
    \includegraphics[trim=2000 15 1500 100, clip, scale=0.08]{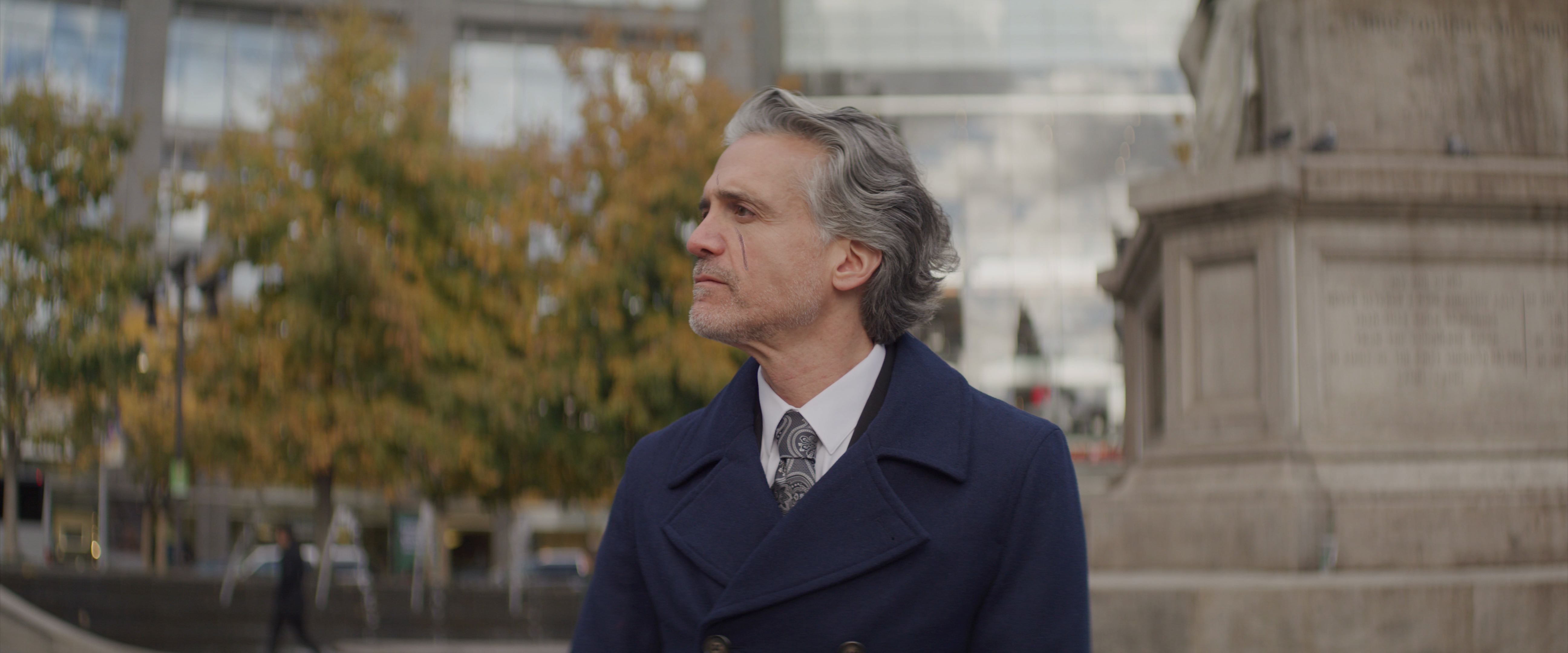}   
  } 
  \subfigure{%
    \includegraphics[trim=2000 15 1500 100, clip, scale=0.08]{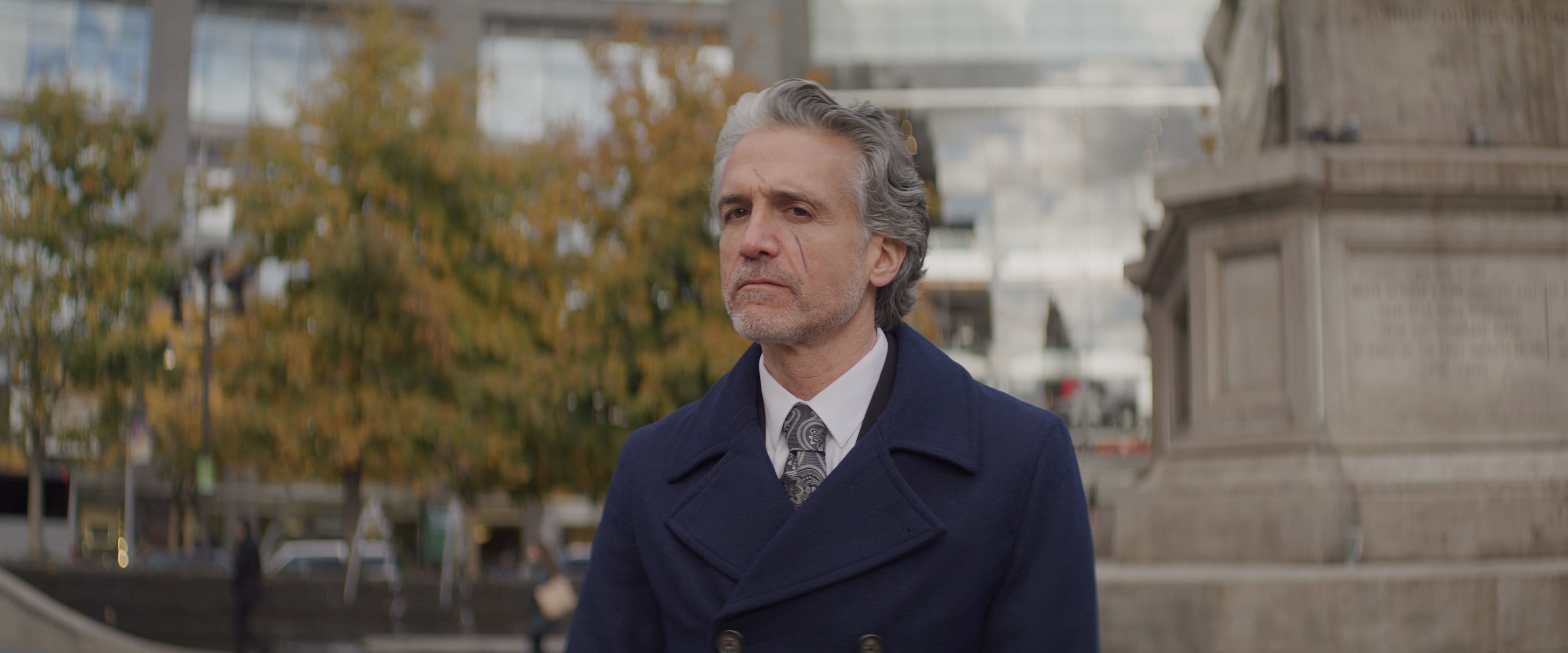} 
  } 
  \subfigure{%
    \includegraphics[trim=2000 15 1500 100, clip, scale=0.08]{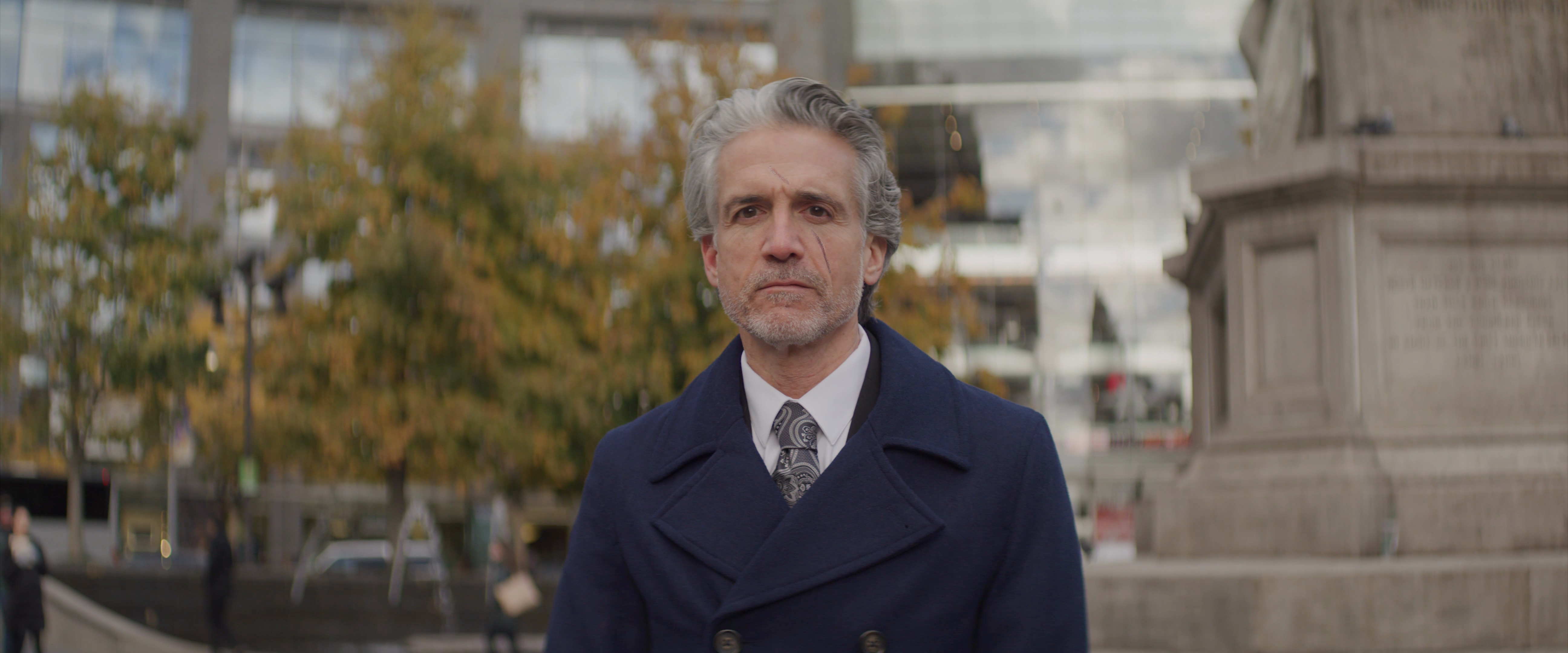}  
  } 
 \caption{Single shot video editing. The artist added the scar on the left cheek and forehead of the actor in the mosaic. 
    The edits are projected back to the frames. We present here the edited frames 0, 50, and 100 (from left to right). The corresponding original frames are presented in figure~\ref{fig:org_seq}.}
    \label{fig:singleshot-editing}
\end{figure*}

We present in figure~\ref{fig:singleshot-editing} an example of video editing using \filmroll in the sequence \textit{FP00673MD02} of the filmpac website. The edition is done as follows: the artist edits the mosaic (here, by adding a scar on the actor's left cheek and another one on the forehead); we compute the difference between the original and edited mosaics; we project and add this difference in all frames of the sequence to obtain the edited video. The spatio-temporal consistency of the edits is very accurate, without any drift across the frames. 
We recall that the corresponding original frames are shown in figure ~\ref{fig:org_seq}.

\subsection{Multiple shot video editing}

\begin{figure*}[ht] 
  \subfigure{%
    \includegraphics[trim=150 75 250 10, clip, scale=0.4]{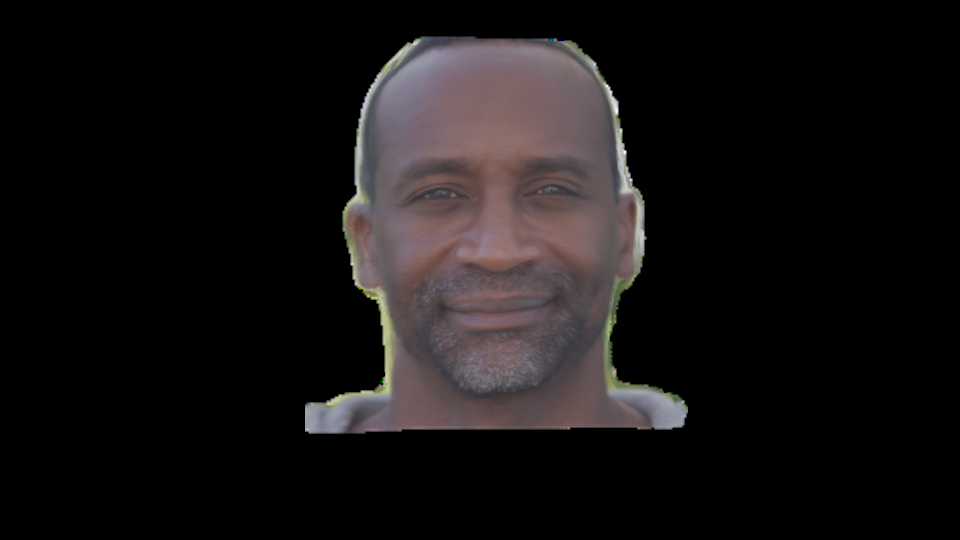} 
  }
  \subfigure{%
    \includegraphics[trim=150 75 250 10, clip, scale=0.4]{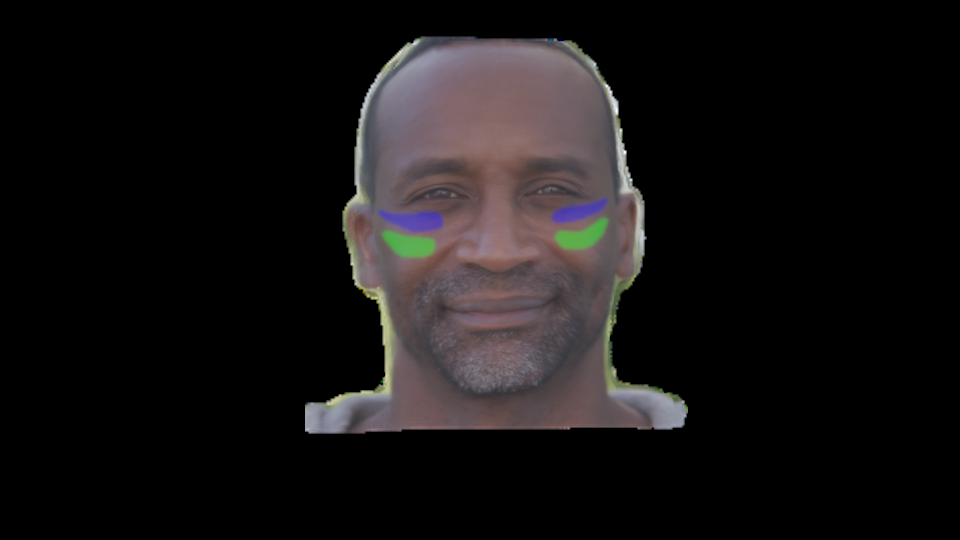}
  }
  \\
  \subfigure{%
    \includegraphics[trim=150 75 250 10, clip, scale=0.4]{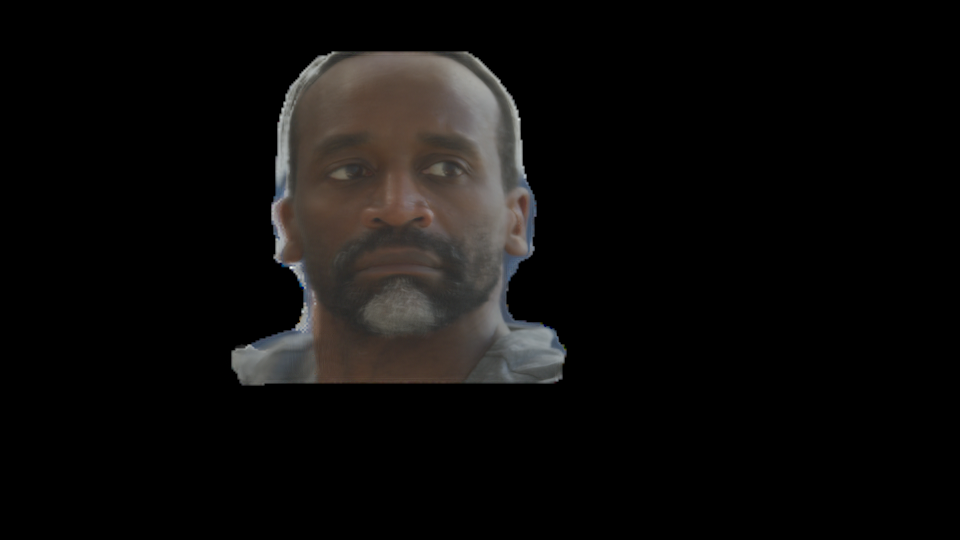}
  }
  \subfigure{%
    \includegraphics[trim=150 75 250 10, clip, scale=0.4]{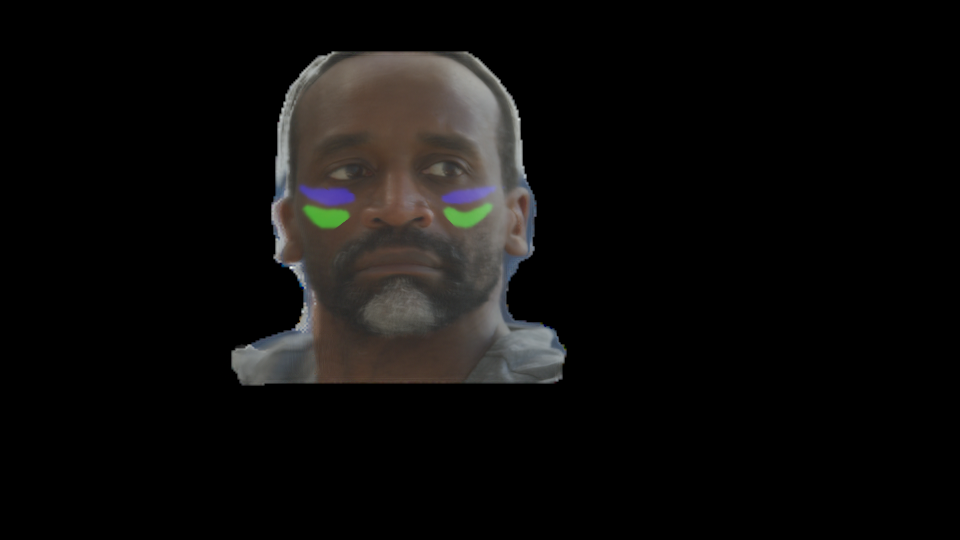}
  }
  \\
   \subfigure{%
    \includegraphics[trim=150 7 150 10, clip, scale=0.22]{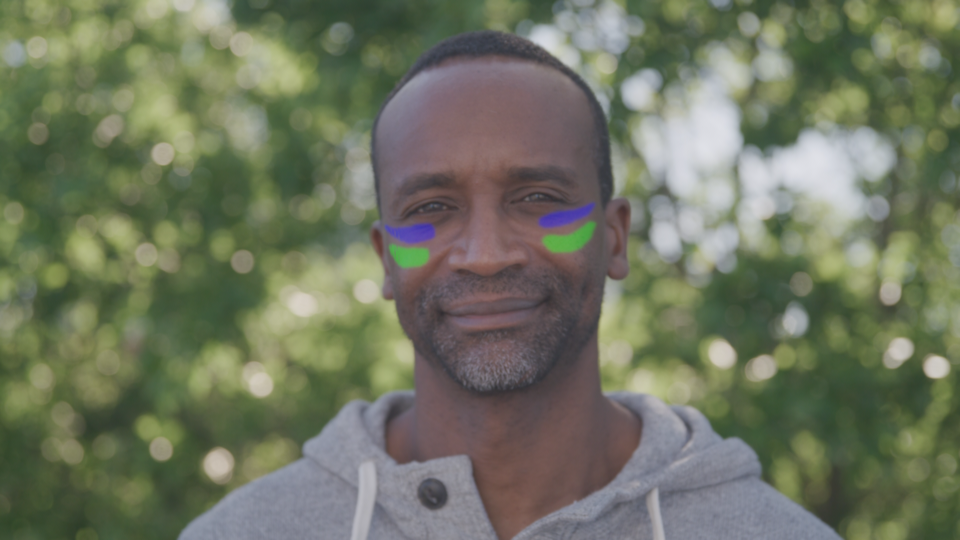}
    }
     \subfigure{%
    \includegraphics[trim=150 7 150 10, clip, scale=0.22]{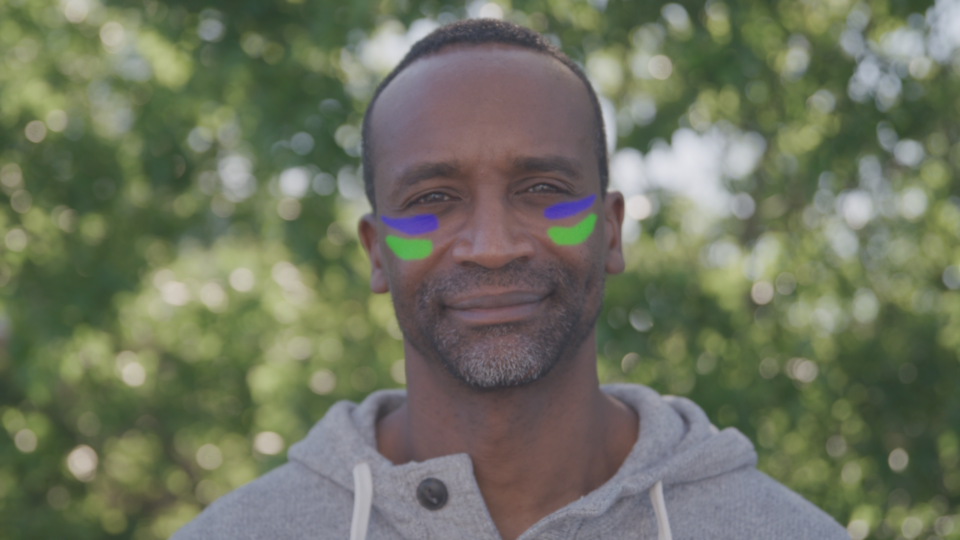}
    }
     \subfigure{%
    \includegraphics[trim=150 7 150 10, clip, scale=0.22]{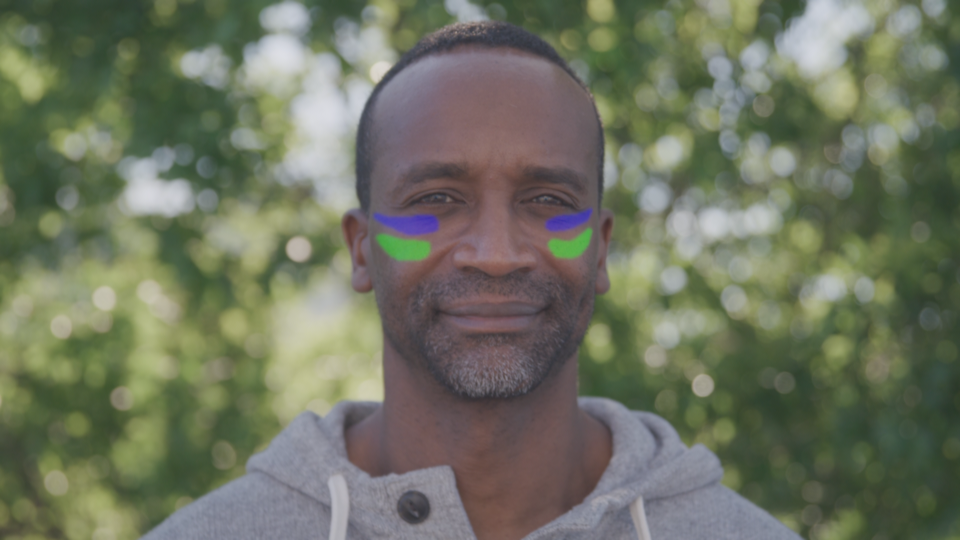}
    }
    \\
\subfigure{%
    \includegraphics[trim=150 7 150 10, clip, scale=0.22]{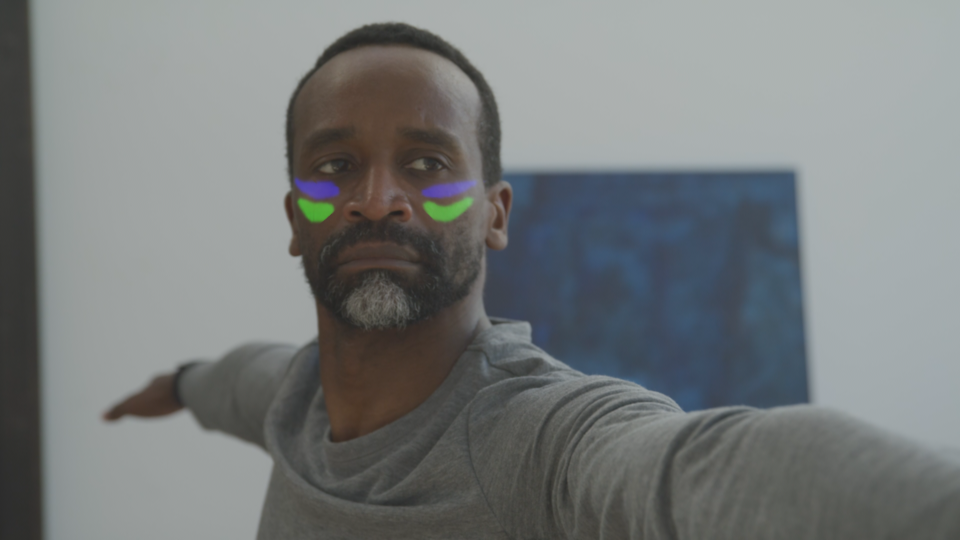}
    }
     \subfigure{%
    \includegraphics[trim=150 7 150 10, clip, scale=0.22]{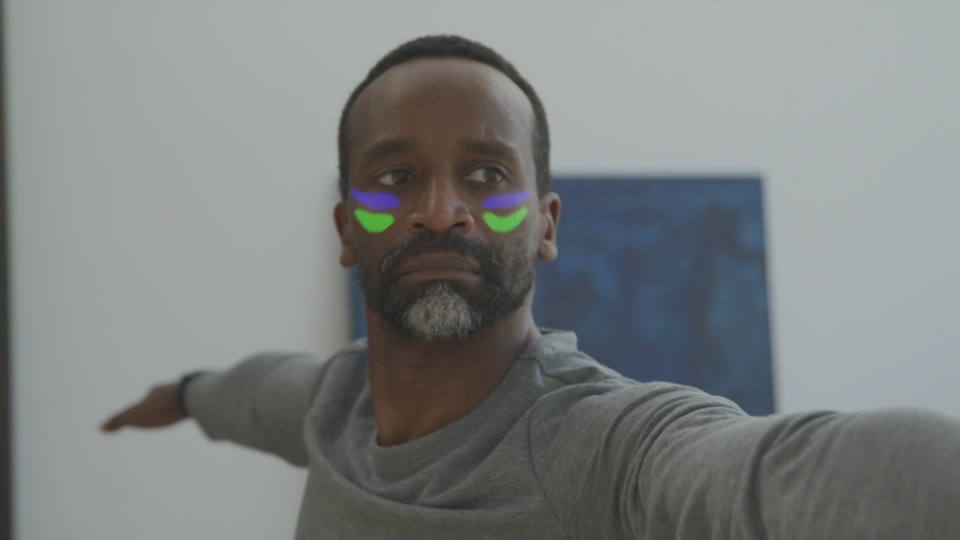}
    }
    \subfigure{%
    \includegraphics[trim=150 7 150 10, clip, scale=0.22]{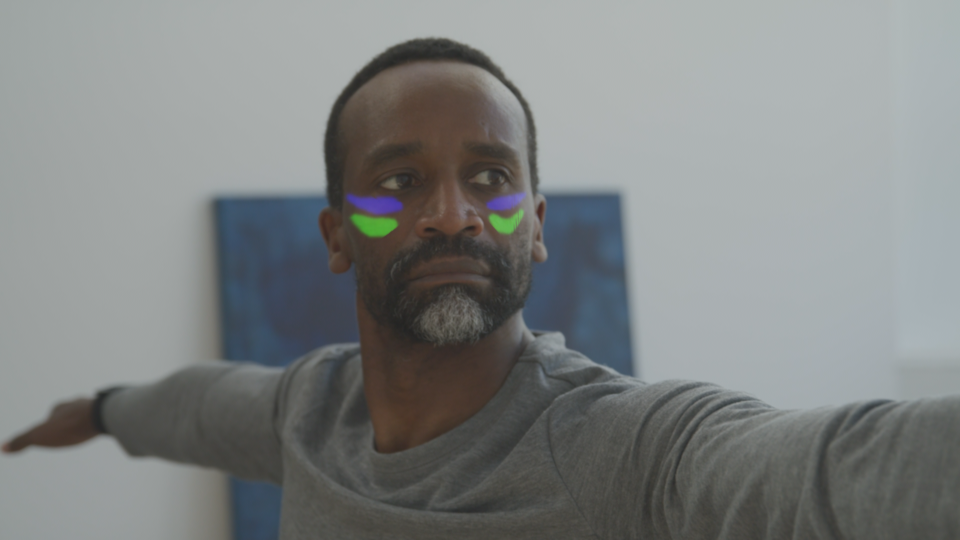}
    }
  \caption{Multishot video editing. Two different shots with same actor are processed to obtain the corresponding mosaics (first column, first and second row). A flag is added on the cheeks of the actor in the first mosaic (top row, second column). The edit is projected automatically in the second mosaic (second row, second column). The edits are finally projected automatically in the corresponding shots (last two rows).}
    \label{fig:multishot-mosaic}
  
  \end{figure*}

For the multi-shot scenario, we select two shots from the filmpac website where the same actor appears: \textit{FP009972MD06}, \textit{FP008841 MD10}. The sequences consist of $153$ and $173$ frames at $4K$ resolution. We first compute the mosaic for each individual shot and then compute the common embedding space for both mosaics. 
{Note that it is not necessary to compute the universal mosaic for editing as we can make use of the bi-directional mapping from one mosaic to another to perform edits on one mosaic and to transfer them to the other mosaics automatically.} The schematic diagram of the multi-shot editing is presented in figure~\ref{fig:multishot-scheme}.
Figure~\ref{fig:multishot-mosaic} shows the individual mosaics and the edited mosaics, where the artists added a flag on both cheeks in the first mosaic (first row). We then automatically transferred the edit on the second mosaic automatically (second row) then re-projected the edit in each frame of each shot.

%
\section{Conclusion} \label{sec:con}
We presented \filmroll, a method that helps post-production artists to edit videos, e.g., for digital makeup. Our method is built upon \textit{unwarp mosaic} \cite{rav-acha2008unwrap}. We proposed the use of a hybrid embedding between two types of tracks. The first type of tracks are obtained using the optical flow between successive frames and permits us to capture the motion of fine details on the face of the actor. The second type of tracks are obtained using a face capture method and permits to capture long range correspondences. We also presented an extension of our method for the multi-shot video editing. Our method was successfully integrated in the workflow of a post-production facility. Our experimental results highlight the potential of \filmroll for editing.  
\bibliographystyle{ACM-Reference-Format}
\bibliography{sample-base}


\begin{thebibliography}{23}


\ifx \showCODEN    \undefined \def \showCODEN     #1{\unskip}     \fi
\ifx \showDOI      \undefined \def \showDOI       #1{#1}\fi
\ifx \showISBNx    \undefined \def \showISBNx     #1{\unskip}     \fi
\ifx \showISBNxiii \undefined \def \showISBNxiii  #1{\unskip}     \fi
\ifx \showISSN     \undefined \def \showISSN      #1{\unskip}     \fi
\ifx \showLCCN     \undefined \def \showLCCN      #1{\unskip}     \fi
\ifx \shownote     \undefined \def \shownote      #1{#1}          \fi
\ifx \showarticletitle \undefined \def \showarticletitle #1{#1}   \fi
\ifx \showURL      \undefined \def \showURL       {\relax}        \fi
\providecommand\bibfield[2]{#2}
\providecommand\bibinfo[2]{#2}
\providecommand\natexlab[1]{#1}
\providecommand\showeprint[2][]{arXiv:#2}

\bibitem[\protect\citeauthoryear{Andrus, Ahn, Alessi, Dib, Gosselin,
  Th\'{e}bault, Chevallier, and Romeo}{Andrus et~al\mbox{.}}{2020}]%
        {andrus2020facelab}
\bibfield{author}{\bibinfo{person}{Curtis Andrus}, \bibinfo{person}{Junghyun
  Ahn}, \bibinfo{person}{Michele Alessi}, \bibinfo{person}{Abdallah Dib},
  \bibinfo{person}{Philippe Gosselin}, \bibinfo{person}{C\'{e}dric
  Th\'{e}bault}, \bibinfo{person}{Louis Chevallier}, {and}
  \bibinfo{person}{Marco Romeo}.} \bibinfo{year}{2020}\natexlab{}.
\newblock \showarticletitle{FaceLab: Scalable Facial Performance Capture for
  Visual Effects}. In \bibinfo{booktitle}{\emph{The Digital Production
  Symposium}} (Virtual Event, USA) \emph{(\bibinfo{series}{DigiPro '20})}.
  \bibinfo{publisher}{Association for Computing Machinery},
  \bibinfo{address}{New York, NY, USA}, Article \bibinfo{articleno}{6},
  \bibinfo{numpages}{3}~pages.
\newblock
\showISBNx{9781450380348}
\urldef\tempurl%
\url{https://doi.org/10.1145/3403736.3403938}
\showDOI{\tempurl}


\bibitem[\protect\citeauthoryear{Beck and Teboulle}{Beck and Teboulle}{2009}]%
        {fista}
\bibfield{author}{\bibinfo{person}{Amir Beck} {and} \bibinfo{person}{Marc
  Teboulle}.} \bibinfo{year}{2009}\natexlab{}.
\newblock \showarticletitle{A Fast Iterative Shrinkage-Thresholding Algorithm
  for Linear Inverse Problems}.
\newblock \bibinfo{journal}{\emph{SIAM J. IMAGING SCIENCES}}
  \bibinfo{volume}{2}, \bibinfo{number}{1} (\bibinfo{year}{2009}),
  \bibinfo{pages}{183--202}.
\newblock


\bibitem[\protect\citeauthoryear{Blanz and Vetter}{Blanz and Vetter}{1999}]%
        {blanz1999mm}
\bibfield{author}{\bibinfo{person}{Volker Blanz} {and} \bibinfo{person}{Thomas
  Vetter}.} \bibinfo{year}{1999}\natexlab{}.
\newblock \showarticletitle{A Morphable Model for the Synthesis of 3D Faces}.
  In \bibinfo{booktitle}{\emph{Proceedings of the 26th Annual Conference on
  Computer Graphics and Interactive Techniques}}
  \emph{(\bibinfo{series}{SIGGRAPH '99})}. \bibinfo{publisher}{ACM
  Press/Addison-Wesley Publishing Co.}, \bibinfo{address}{USA},
  \bibinfo{pages}{187–194}.
\newblock
\showISBNx{0201485605}
\urldef\tempurl%
\url{https://doi.org/10.1145/311535.311556}
\showDOI{\tempurl}


\bibitem[\protect\citeauthoryear{Bulat and Tzimiropoulos}{Bulat and
  Tzimiropoulos}{2017}]%
        {bulat17far}
\bibfield{author}{\bibinfo{person}{Adrian Bulat} {and}
  \bibinfo{person}{Georgios Tzimiropoulos}.} \bibinfo{year}{2017}\natexlab{}.
\newblock \showarticletitle{How far are we from solving the 2D \& 3D Face
  Alignment problem? (and a dataset of 230,000 3D facial landmarks)}. In
  \bibinfo{booktitle}{\emph{International Conference on Computer Vision}}.
\newblock


\bibitem[\protect\citeauthoryear{Dib, Bharaj, Ahn, Th{\'{e}}bault, Gosselin,
  Romeo, and Chevallier}{Dib et~al\mbox{.}}{2021}]%
        {dib2021drt}
\bibfield{author}{\bibinfo{person}{Abdallah Dib}, \bibinfo{person}{Gaurav
  Bharaj}, \bibinfo{person}{Junghyun Ahn}, \bibinfo{person}{C{\'{e}}dric
  Th{\'{e}}bault}, \bibinfo{person}{Philippe~Henri Gosselin},
  \bibinfo{person}{Marco Romeo}, {and} \bibinfo{person}{Louis Chevallier}.}
  \bibinfo{year}{2021}\natexlab{}.
\newblock \showarticletitle{Practical Face Reconstruction via Differentiable
  Ray Tracing}.
\newblock \bibinfo{journal}{\emph{Comput. Graph. Forum}} \bibinfo{volume}{40},
  \bibinfo{number}{2} (\bibinfo{year}{2021}), \bibinfo{pages}{153--164}.
\newblock
\urldef\tempurl%
\url{https://doi.org/10.1111/cgf.142622}
\showDOI{\tempurl}


\bibitem[\protect\citeauthoryear{Egger, Smith, Tewari, Wuhrer, Zollhoefer,
  Beeler, Bernard, Bolkart, Kortylewski, Romdhani, Theobalt, Blanz, and
  Vetter}{Egger et~al\mbox{.}}{2020}]%
        {egger2020mmfuture}
\bibfield{author}{\bibinfo{person}{Bernhard Egger}, \bibinfo{person}{William
  A.~P. Smith}, \bibinfo{person}{Ayush Tewari}, \bibinfo{person}{Stefanie
  Wuhrer}, \bibinfo{person}{Michael Zollhoefer}, \bibinfo{person}{Thabo
  Beeler}, \bibinfo{person}{Florian Bernard}, \bibinfo{person}{Timo Bolkart},
  \bibinfo{person}{Adam Kortylewski}, \bibinfo{person}{Sami Romdhani},
  \bibinfo{person}{Christian Theobalt}, \bibinfo{person}{Volker Blanz}, {and}
  \bibinfo{person}{Thomas Vetter}.} \bibinfo{year}{2020}\natexlab{}.
\newblock \showarticletitle{3D Morphable Face Models—Past, Present, and
  Future}.
\newblock \bibinfo{journal}{\emph{ACM Trans. Graph.}} \bibinfo{volume}{39},
  \bibinfo{number}{5}, Article \bibinfo{articleno}{157} (\bibinfo{date}{June}
  \bibinfo{year}{2020}), \bibinfo{numpages}{38}~pages.
\newblock
\showISSN{0730-0301}
\urldef\tempurl%
\url{https://doi.org/10.1145/3395208}
\showDOI{\tempurl}


\bibitem[\protect\citeauthoryear{Fortun, Bouthemy, and Kervrann}{Fortun
  et~al\mbox{.}}{2015}]%
        {fortun2015optical}
\bibfield{author}{\bibinfo{person}{Denis Fortun}, \bibinfo{person}{Patrick
  Bouthemy}, {and} \bibinfo{person}{Charles Kervrann}.}
  \bibinfo{year}{2015}\natexlab{}.
\newblock \showarticletitle{Optical flow modeling and computation: A survey}.
\newblock \bibinfo{journal}{\emph{Computer Vision and Image Understanding}}
  \bibinfo{volume}{134} (\bibinfo{year}{2015}), \bibinfo{pages}{1--21}.
\newblock


\bibitem[\protect\citeauthoryear{Garrido, Zollh\"{o}fer, Casas, Valgaerts,
  Varanasi, P\'{e}rez, and Theobalt}{Garrido et~al\mbox{.}}{2016}]%
        {garrido2016mono}
\bibfield{author}{\bibinfo{person}{Pablo Garrido}, \bibinfo{person}{Michael
  Zollh\"{o}fer}, \bibinfo{person}{Dan Casas}, \bibinfo{person}{Levi
  Valgaerts}, \bibinfo{person}{Kiran Varanasi}, \bibinfo{person}{Patrick
  P\'{e}rez}, {and} \bibinfo{person}{Christian Theobalt}.}
  \bibinfo{year}{2016}\natexlab{}.
\newblock \showarticletitle{Reconstruction of Personalized 3D Face Rigs from
  Monocular Video}.
\newblock \bibinfo{journal}{\emph{ACM Trans. Graph.}} \bibinfo{volume}{35},
  \bibinfo{number}{3}, Article \bibinfo{articleno}{28} (\bibinfo{date}{May}
  \bibinfo{year}{2016}), \bibinfo{numpages}{15}~pages.
\newblock
\showISSN{0730-0301}
\urldef\tempurl%
\url{https://doi.org/10.1145/2890493}
\showDOI{\tempurl}


\bibitem[\protect\citeauthoryear{Grover and Leskovec}{Grover and
  Leskovec}{2016}]%
        {Adityanode2vec}
\bibfield{author}{\bibinfo{person}{Aditya Grover} {and} \bibinfo{person}{Jure
  Leskovec}.} \bibinfo{year}{2016}\natexlab{}.
\newblock \showarticletitle{Node2vec: Scalable Feature Learning for Networks}.
  In \bibinfo{booktitle}{\emph{Proceedings of the 22nd ACM SIGKDD International
  Conference on Knowledge Discovery and Data Mining}} (San Francisco,
  California, USA) \emph{(\bibinfo{series}{KDD '16})}.
  \bibinfo{publisher}{Association for Computing Machinery},
  \bibinfo{address}{New York, NY, USA}, \bibinfo{pages}{855–864}.
\newblock
\showISBNx{9781450342322}
\urldef\tempurl%
\url{https://doi.org/10.1145/2939672.2939754}
\showDOI{\tempurl}


\bibitem[\protect\citeauthoryear{Kingma and Ba}{Kingma and Ba}{2017}]%
        {kingma2017adam}
\bibfield{author}{\bibinfo{person}{Diederik~P. Kingma} {and}
  \bibinfo{person}{Jimmy Ba}.} \bibinfo{year}{2017}\natexlab{}.
\newblock \bibinfo{title}{Adam: A Method for Stochastic Optimization}.
\newblock
\newblock
\showeprint[arxiv]{1412.6980}~[cs.LG]


\bibitem[\protect\citeauthoryear{Kirillov, Wu, He, and Girshick}{Kirillov
  et~al\mbox{.}}{2020}]%
        {kirillov2020pointrend}
\bibfield{author}{\bibinfo{person}{Alexander Kirillov}, \bibinfo{person}{Yuxin
  Wu}, \bibinfo{person}{Kaiming He}, {and} \bibinfo{person}{Ross Girshick}.}
  \bibinfo{year}{2020}\natexlab{}.
\newblock \showarticletitle{Pointrend: Image segmentation as rendering}. In
  \bibinfo{booktitle}{\emph{Proceedings of the IEEE/CVF conference on computer
  vision and pattern recognition}}. \bibinfo{pages}{9799--9808}.
\newblock


\bibitem[\protect\citeauthoryear{Marvasti-Zadeh, Cheng, Ghanei-Yakhdan, and
  Kasaei}{Marvasti-Zadeh et~al\mbox{.}}{2021}]%
        {marvasti2021deep}
\bibfield{author}{\bibinfo{person}{Seyed~Mojtaba Marvasti-Zadeh},
  \bibinfo{person}{Li Cheng}, \bibinfo{person}{Hossein Ghanei-Yakhdan}, {and}
  \bibinfo{person}{Shohreh Kasaei}.} \bibinfo{year}{2021}\natexlab{}.
\newblock \showarticletitle{Deep learning for visual tracking: A comprehensive
  survey}.
\newblock \bibinfo{journal}{\emph{IEEE Transactions on Intelligent
  Transportation Systems}} (\bibinfo{year}{2021}).
\newblock


\bibitem[\protect\citeauthoryear{McInnes, Healy, and Melville}{McInnes
  et~al\mbox{.}}{2020}]%
        {mcinnes2020umap}
\bibfield{author}{\bibinfo{person}{Leland McInnes}, \bibinfo{person}{John
  Healy}, {and} \bibinfo{person}{James Melville}.}
  \bibinfo{year}{2020}\natexlab{}.
\newblock \bibinfo{title}{UMAP: Uniform Manifold Approximation and Projection
  for Dimension Reduction}.
\newblock
\newblock
\showeprint[arxiv]{1802.03426}~[stat.ML]


\bibitem[\protect\citeauthoryear{Ng, Jordan, and Weiss}{Ng
  et~al\mbox{.}}{2001}]%
        {Ngspectralemb}
\bibfield{author}{\bibinfo{person}{Andrew~Y. Ng}, \bibinfo{person}{Michael~I.
  Jordan}, {and} \bibinfo{person}{Yair Weiss}.}
  \bibinfo{year}{2001}\natexlab{}.
\newblock \showarticletitle{On Spectral Clustering: Analysis and an Algorithm}.
  In \bibinfo{booktitle}{\emph{Proceedings of the 14th International Conference
  on Neural Information Processing Systems: Natural and Synthetic}} (Vancouver,
  British Columbia, Canada) \emph{(\bibinfo{series}{NIPS'01})}.
  \bibinfo{publisher}{MIT Press}, \bibinfo{address}{Cambridge, MA, USA},
  \bibinfo{pages}{849–856}.
\newblock


\bibitem[\protect\citeauthoryear{Rav-Acha, Kohli, Rother, and
  Fitzgibbon}{Rav-Acha et~al\mbox{.}}{2008}]%
        {rav-acha2008unwrap}
\bibfield{author}{\bibinfo{person}{Alex Rav-Acha}, \bibinfo{person}{Pushmeet
  Kohli}, \bibinfo{person}{Carsten Rother}, {and} \bibinfo{person}{Andrew
  Fitzgibbon}.} \bibinfo{year}{2008}\natexlab{}.
\newblock \showarticletitle{Unwrap Mosaics: A New Representation for Video
  Editing}. In \bibinfo{booktitle}{\emph{SIGGRAPH '08 ACM SIGGRAPH 2008 papers}
  (\bibinfo{edition}{siggraph '08 acm siggraph 2008 papers} ed.)}.
  \bibinfo{publisher}{ACM}.
\newblock
\urldef\tempurl%
\url{https://www.microsoft.com/en-us/research/publication/unwrap-mosaics-new-representation-video-editing/}
\showURL{%
\tempurl}


\bibitem[\protect\citeauthoryear{Rocco, Arandjelovi{\'c}, and Sivic}{Rocco
  et~al\mbox{.}}{2018}]%
        {rocco2018end}
\bibfield{author}{\bibinfo{person}{Ignacio Rocco}, \bibinfo{person}{Relja
  Arandjelovi{\'c}}, {and} \bibinfo{person}{Josef Sivic}.}
  \bibinfo{year}{2018}\natexlab{}.
\newblock \showarticletitle{End-to-end weakly-supervised semantic alignment}.
  In \bibinfo{booktitle}{\emph{Proceedings of the IEEE Conference on Computer
  Vision and Pattern Recognition}}. \bibinfo{pages}{6917--6925}.
\newblock


\bibitem[\protect\citeauthoryear{Roweis and Saul}{Roweis and Saul}{2000}]%
        {Roweis2323LLE}
\bibfield{author}{\bibinfo{person}{Sam~T. Roweis} {and}
  \bibinfo{person}{Lawrence~K. Saul}.} \bibinfo{year}{2000}\natexlab{}.
\newblock \showarticletitle{Nonlinear Dimensionality Reduction by Locally
  Linear Embedding}.
\newblock \bibinfo{journal}{\emph{Science}} \bibinfo{volume}{290},
  \bibinfo{number}{5500} (\bibinfo{year}{2000}), \bibinfo{pages}{2323--2326}.
\newblock
\showISSN{0036-8075}
\urldef\tempurl%
\url{https://doi.org/10.1126/science.290.5500.2323}
\showDOI{\tempurl}
\showeprint{https://science.sciencemag.org/content/290/5500/2323.full.pdf}


\bibitem[\protect\citeauthoryear{Smeulders, Chu, Cucchiara, Calderara, Dehghan,
  and Shah}{Smeulders et~al\mbox{.}}{2013}]%
        {smeulders2013visual}
\bibfield{author}{\bibinfo{person}{Arnold~WM Smeulders},
  \bibinfo{person}{Dung~M Chu}, \bibinfo{person}{Rita Cucchiara},
  \bibinfo{person}{Simone Calderara}, \bibinfo{person}{Afshin Dehghan}, {and}
  \bibinfo{person}{Mubarak Shah}.} \bibinfo{year}{2013}\natexlab{}.
\newblock \showarticletitle{Visual tracking: An experimental survey}.
\newblock \bibinfo{journal}{\emph{IEEE transactions on pattern analysis and
  machine intelligence}} \bibinfo{volume}{36}, \bibinfo{number}{7}
  (\bibinfo{year}{2013}), \bibinfo{pages}{1442--1468}.
\newblock


\bibitem[\protect\citeauthoryear{van~der Maaten and Hinton}{van~der Maaten and
  Hinton}{2008}]%
        {Maatentsne}
\bibfield{author}{\bibinfo{person}{Laurens van~der Maaten} {and}
  \bibinfo{person}{Geoffrey Hinton}.} \bibinfo{year}{2008}\natexlab{}.
\newblock \showarticletitle{Visualizing Data using t-SNE}.
\newblock \bibinfo{journal}{\emph{Journal of Machine Learning Research}}
  \bibinfo{volume}{9}, \bibinfo{number}{86} (\bibinfo{year}{2008}),
  \bibinfo{pages}{2579--2605}.
\newblock


\bibitem[\protect\citeauthoryear{Weinzaepfel, Revaud, Harchaoui, and
  Schmid}{Weinzaepfel et~al\mbox{.}}{2013}]%
        {Weinzaepfeldeepflow}
\bibfield{author}{\bibinfo{person}{P. Weinzaepfel}, \bibinfo{person}{J.
  Revaud}, \bibinfo{person}{Z. Harchaoui}, {and} \bibinfo{person}{C. Schmid}.}
  \bibinfo{year}{2013}\natexlab{}.
\newblock \showarticletitle{DeepFlow: Large Displacement Optical Flow with Deep
  Matching}. In \bibinfo{booktitle}{\emph{2013 IEEE International Conference on
  Computer Vision (ICCV)}}. \bibinfo{publisher}{IEEE Computer Society},
  \bibinfo{address}{Los Alamitos, CA, USA}, \bibinfo{pages}{1385--1392}.
\newblock
\showISSN{1550-5499}
\urldef\tempurl%
\url{https://doi.org/10.1109/ICCV.2013.175}
\showDOI{\tempurl}


\bibitem[\protect\citeauthoryear{Wu, Kirillov, Massa, Lo, and Girshick}{Wu
  et~al\mbox{.}}{2019}]%
        {wu2019detectron2}
\bibfield{author}{\bibinfo{person}{Yuxin Wu}, \bibinfo{person}{Alexander
  Kirillov}, \bibinfo{person}{Francisco Massa}, \bibinfo{person}{Wan-Yen Lo},
  {and} \bibinfo{person}{Ross Girshick}.} \bibinfo{year}{2019}\natexlab{}.
\newblock \bibinfo{title}{Detectron2}.
\newblock
  \bibinfo{howpublished}{\url{https://github.com/facebookresearch/detectron2}}.
\newblock


\bibitem[\protect\citeauthoryear{Yilmaz, Javed, and Shah}{Yilmaz
  et~al\mbox{.}}{2006}]%
        {yilmaz2006object}
\bibfield{author}{\bibinfo{person}{Alper Yilmaz}, \bibinfo{person}{Omar Javed},
  {and} \bibinfo{person}{Mubarak Shah}.} \bibinfo{year}{2006}\natexlab{}.
\newblock \showarticletitle{Object tracking: A survey}.
\newblock \bibinfo{journal}{\emph{Acm computing surveys (CSUR)}}
  \bibinfo{volume}{38}, \bibinfo{number}{4} (\bibinfo{year}{2006}),
  \bibinfo{pages}{13--es}.
\newblock


\bibitem[\protect\citeauthoryear{Zhang, Zhu, Lei, Shi, Wang, and Li}{Zhang
  et~al\mbox{.}}{2017}]%
        {zhang17iccv}
\bibfield{author}{\bibinfo{person}{Shifeng Zhang}, \bibinfo{person}{Xiangyu
  Zhu}, \bibinfo{person}{Zhen Lei}, \bibinfo{person}{Hailin Shi},
  \bibinfo{person}{Xiaobo Wang}, {and} \bibinfo{person}{Stan~Z. Li}.}
  \bibinfo{year}{2017}\natexlab{}.
\newblock \showarticletitle{{S$^3$FD}: Single Shot Scale Invariant Face
  Detector}. In \bibinfo{booktitle}{\emph{IEEE International Conference on
  Computer Vision (ICCV)}}.
\newblock


\end{thebibliography}

\appendix

\end{document}